# Soft Sensor for Bottom-Hole Pressure Estimation in Petroleum Wells Using LSTM and Transfer Learning

M. A. Fernandes[a,b,*], E. Gildin[c] and M. A. Sampaio[a]

[a]Department of Mining and Petroleum Engineering, University of São Paulo, São Paulo, SP, Brazil
[b]Department of Reservoir and Production Engineering, Petrobras, Santos, SP, Brazil
[c]Department of Petroleum Engineering, Texas A&M University, College Station, TX, USA

ABSTRACT

Monitoring bottom-hole variables in petroleum wells is essential for production optimization, asset integrity, and emissions reduction. Permanent Downhole Gauges (PDGs) provide real-time pressure data but face high failure rates and prohibitive replacement costs in harsh downhole environments. This paper develops a data-driven soft sensor to estimate flowing Bottom-Hole Pressure (BHP) using topside and wellhead measurements. A Long Short-Term Memory (LSTM) network is introduced to capture temporal dependencies in multiphase flow conditions and compared against Multi-Layer Perceptron (MLP) and Ridge Regression benchmarks. We further implement a Transfer Learning framework that adapts pre-trained models across oilfields with different operational characteristics without complete retraining, addressing a critical scalability challenge for AI deployment in petroleum engineering. The domain shift between fields is formally quantified, providing a physically interpretable justification for the transfer learning framework. Validated on real offshore datasets from Brazil's ultra-deepwater Pre-salt basin (60 wells, 13 years, over 100,000 samples), the methodology achieved MAPE consistently below 2%, outperforming existing literature benchmarks. A stratified error analysis further demonstrates that the LSTM presents a consistent advantage over simpler models in atypical operating conditions, precisely where accurate BHP estimation is most critical. The proposed framework supports digital twin applications in smart oilfields, offering a scalable and cost-effective alternative to physical sensors under diverse reservoir and flow conditions.

## 1. Introduction

One of the primary challenges in petroleum engineering is dealing with the lack of direct access to the source of natural resources: the subsurface reservoir rock. Throughout the exploration and production (E&P) life cycle, this challenge is addressed in various ways. For instance, during the exploration phase, we rely on indirect data acquisition, such as seismic surveys, while in the early development phase, we obtain partial insights through rock samples collected during drilling operations. However, the most enduring source of information throughout the life cycle of wells and reservoirs is their behavior during production. Measurements of pressure, temperature, and production rates over time, although taken sparsely at the well locations, form the foundation for detecting anomalies, early identification of production loss events (e.g., scaling or reservoir damage), understanding reservoir configuration, forecasting future performance, and ultimately guiding actions and projects aimed at optimizing recoverable volumes [1], as well as operational and environmental safety and, ultimately, value addition. The availability of continuous and reliable data for these variables during the productive life of a well is therefore critical for disciplines such as reservoir engineering, well testing, flow assurance, and operations management [2].

In offshore production systems, the most relevant nodes for such measurements are:

- Bottom-hole: where pressures and temperatures are measured by the Permanent Downhole Gauges (PDGs).
- Christmas tree at the wellhead: where the same variables are tracked via temperature and pressure transducers (TPTs).
- Topside of the platform: where we find sensors for pressure and temperature both at the choke valve inlet and outlet (the ones which regulate the flow rate and pressure for each well), as well as flowmeters and monitoring systems for oil, gas, and water rates. A schematic of an offshore production system, highlighting its basic components and the aforementioned analysis nodes is shown in Figure 1 [3].

The PDGs, positioned at the wellbore's bottom-hole and closest to the reservoir, serve as the most critical sensors for reservoir management and production monitoring [4]. While these sensors are widely employed in high-yield offshore wells, the deployment and maintenance expenses for the instruments themselves, along with their accompanying monitoring infrastructure, often render them economically unfeasible in many scenarios. This is particularly evident in mature fields and in cases where low productivity or short well lifespan is anticipated. As a result, in practice, PDGs are installed in only a minority of wells across the industry.

When PDG data are inaccessible, obtaining bottom-hole pressure measurements typically relies on periodic wireline interventions. Those operations require interruptions in production to deploy downhole testing equipment. Even in wells originally equipped with PDGs, data accessibility may be compromised over time due to gauge malfunctions or communication breakdowns with topside monitoring systems [5]. Operating for extended periods in high-pressure, inaccessible downhole environments, these sensors have the highest failure risk, and their replacement demands complex workover operations that are generally economically prohibitive [6].

*Corresponding author
matfernan@petrobras.com.br (M.A. Fernandes)
ORCID(s): 0000-0002-6416-5061 (M.A. Fernandes); 0000-0001-5985-4500 (E. Gildin); 0000-0003-1125-7218 (M.A. Sampaio)

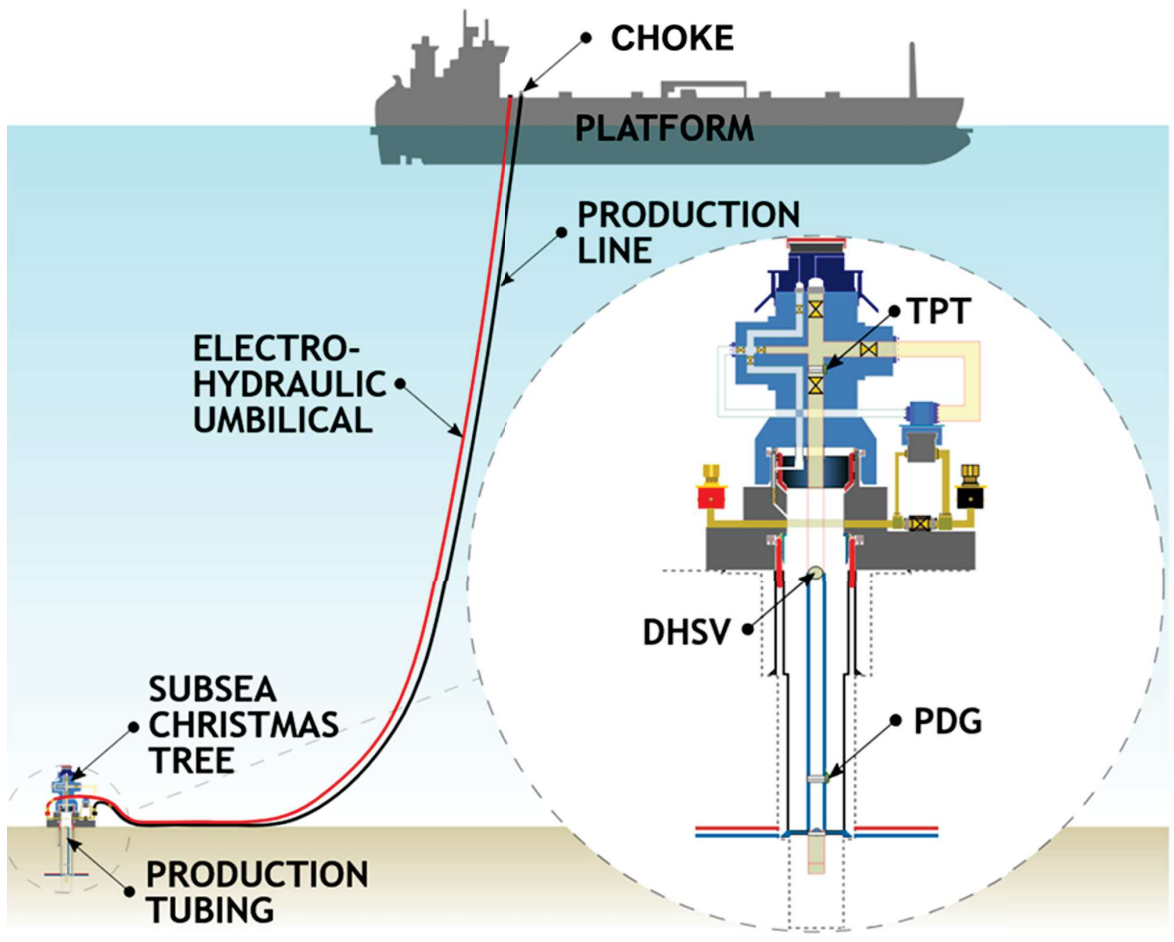

**Figure 1:** Schematic of an offshore production system, highlighting the positioning of the most relevant equipment and sensors. Reproduced from [3].

An alternative for unavailable sensor conditions is virtual metering (soft sensors). This approach can be based on techniques as machine learning, mathematical relationships, statistical methods, and potential integrations with other tools to monitor process variables when physical measures (hard sensors) cannot be used, including hostile environments, information delays, and others [7].

This work focuses on developing a data-driven soft sensor for flowing bottom-hole pressure estimation. Here we emphasize that the proposed applicability is limited to steady-state flow pressure. This pressure is lower than the reservoir's static pressure, resulting in the pressure differential required for fluid flow from the porous medium to the wellbore, as described by Darcy's Law [8].

This alternative is designed to address scenarios where physical downhole gauges are unavailable or to serve as a component of a digital twin system (referring to a virtual representation of equipment or processes) [9]. In such cases, the machine learning-estimated BHP can be continuously compared to the actual measurements, providing a reference value for error detection and anomaly identification.

The following section brings a contextualization, providing an overview of traditional approaches and state-of-the-art methods for this task, and inserting the proposed solution within current technological gaps. Section 3 presents the proposed methodology, systematically explaining each step of model development. Section 4 brings the results and discusses their practical implications, while Section 5 summarizes key conclusions.

## 2. Contextualization and Objectives

In oil production, we deal with a complex case of multiphase flow through pipelines with varying geometries, resulting in diverse flow patterns. This combination of factors poses significant challenges for modeling, leading the industry to rely primarily on empirical, data-driven, or hybrid solutions. In this section, we detail the problem and highlight some of the most relevant existing solutions, as well as the gap that our work aims to bridge.

### 2.1. Problem Statement

A fundamental characteristic in oil production is the gas liberation due to depressurization, once the bubble-point pressure is reached. This process initiates with the lighter hydrocarbon fractions of the mixture transitioning from liquid to gas phase [10], occurring either in the reservoir after depletion or during its flow toward the surface. The conditions required for bubble point pressure to be reached vary over time, depending on pressure drops along the flow path, restrictions in well openings such as choke valves, reservoir drainage status, and changes in produced fluid composition, which may also change temporally (concept drift) due to reservoir characteristics or injection breakthrough.

The multiphase flow eventually becomes more complex over time, as water production generally increases coming from injection or aquifer influx [8], resulting in three concomitant phases. Flow geometry further complicates the system: production typically involves vertical flow in the well tubing, transitions to horizontal flow in seabed-mounted flowlines (in offshore systems), and resumes vertical flow in risers leading to the platform. Deviations such as downward flow sections may occur in configurations like lazy-wave risers, adding further variability to flow patterns [11].

Other influencing factors include well control valve operations—particularly the choke valve, which regulates production by introducing a pressure drop that affects the entire flow system—as well as intelligent completion equipment (used to modify the open reservoir interval for production) and artificial lift methods. For example, gas lift, a common artificial lift technique, reduces the average fluid density by injecting gas into the wellbore, thereby lowering the flowing BHP [12]. This method is illustrated in Figure 2.

This combination of factors can lead to multiple flow patterns along the wellbore and pipelines, from bottom-hole to the surface, often varying over time depending on well conditions. In vertical pipes, common flow regimes include bubble, slug, churn, annular, and wispy, while horizontal flow typically exhibits annular, bubble, stratified, and intermittent patterns. Pipe orientation significantly influences these regimes: vertical flow maintains axial symmetry, whereas horizontal flow tends toward stratification due to gravitational effects [13].

The interplay of changes in flow patterns, mass transfer between phases, and slippage between phases establishes complex nonlinear relationships in pressure gradients [14],

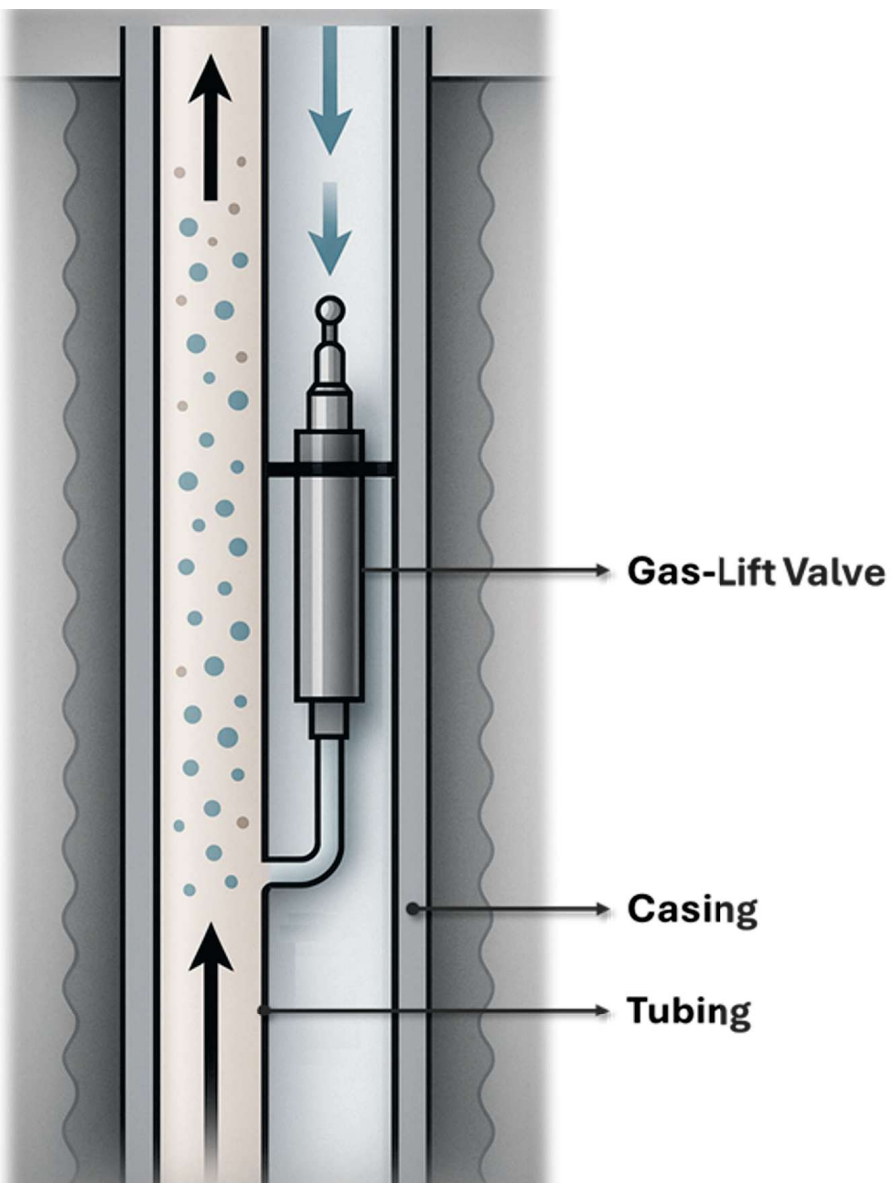

**Figure 2:** Schematic representation of a gas lift valve operating within an oil well. Image enhanced using generative AI tools.

where gravitational, frictional, and accelerational components:

$$\left.\frac{dP}{dL}\right|_{Total} = \left.\frac{dP}{dL}\right|_{grav} + \left.\frac{dP}{dL}\right|_{fric} + \left.\frac{dP}{dL}\right|_{accel} \tag{1}$$

where $dP/dL$ are the pressure drops along the lengths of tubing and pipelines, and each component responds differently to changing flow regimes, fluid compositions, and control actions. This represents a challenge for obtaining precise analytical modeling and even for accurate numerical simulations that are valid for a broad range of conditions [13].

## 2.2. Traditional Methods

Traditional multiphase flow modeling relies heavily on empirical correlations derived from experimental data. Some of the more notable works in this area are listed in the next paragraphs, which continue to be widely used in the industry decades after their publication:

Duns and Ros [15] developed a correlation for estimating pressure gradients in vertical multiphase flow, based on experimental data covering a broad range of oil–gas mixtures with varying water cuts. The method incorporates flow regime maps to identify flow patterns and estimates pressure drops by accounting for liquid holdup, phase slippage, frictional losses, and acceleration effects. The method remains widely adopted due to its accuracy in high gas fraction scenarios, including production under gas lift.

Hagedorn and Brown [16] also worked with vertical flow, based on full-scale experimental data. Their method assumes steady-state, homogeneous flow and incorporates pressure- and temperature-dependent fluid properties, used in iterative liquid holdup calculations. The authors intended to obtain a more general correlation, and for that relied on dimensionless groups and avoided segregation for different flow patterns. This generalization allows accurate predictions across a wide range of tubing sizes, flow conditions, and fluid compositions. However, its accuracy diminishes in deviated or horizontal wells due to increased flow regime complexity.

Beggs and Brill [14] were pioneers in developing a correlation capable of handling inclined pipes at any angle. Their method accounts for different flow regimes (segregated, intermittent, and distributed) and calculates pressure gradients using holdup and friction factor adjustments. It is widely used in the petroleum industry due to its versatility and accuracy in modeling multiphase flow in production tubing and flowlines

These correlations are present even in the foundations of many commercial simulation tools, due to their computational efficiency and proven performance in conventional scenarios.

Despite their widespread use, these correlations tend to be limited by their calibration for particular conditions, which reduces their accuracy when applied to a broader range of production scenarios. As wells and fields evolve over time, changes in flow regimes can lead to reduced reliability, especially as production dynamics shift. Additionally, many empirical models simplify physical phenomena, such as phase slippage or assuming homogeneous fluid, which can lead to inaccuracies under complex conditions [17].

Mechanistic models, while based in flow physics, also face similar challenges while assuming a single flow regime and oversimplifying governing dynamics [18].

As production systems continue to diversify, e.g. high-GOR shale wells or $CO_2$-rich flows, the rigidity of these correlations poses a growing challenge for accurate performance prediction and optimization.

The limitations of application ranges also affect simulation software. While these tools are capable of providing more precise estimates for pressure and temperature gradients, performing sensitivity analyses, managing gas lift operations, and calculating flow rates from boundary pressures, they still depend on models that require periodic updates to parameters reflecting current well conditions, such as inflow performance relationships (IPR) [17].

## 2.3. Current Research

With the advent of data-driven, artificial intelligence, and deep learning methods, this area of research headed to a new direction. Those methods offer an advantage here by enabling the development of soft sensors even when the relationships between variables are not explicitly known, and are increasingly replacing conventional instrumentation for real-time monitoring and automated process control [19]. Developments on analytical and numerical solutions, nonetheless, are still useful for studies of multiphase flow, especially in cases where we must calculate pressure losses in tubing and pipelines [20].

A broad set of works explores data-driven approaches under diverse well types and operating conditions. Campos et al. [21] estimated flowing BHP using radial basis function neural networks optimized through particle swarm optimization. Nwanwe and Duru [22] gave preference to a white-box approach via an adaptive neuro-fuzzy model, which performed significantly better than empirical correlations. Opoku et al. [23] selected empirical correlations through a Quasi-Monte Carlo method. Sanusi et al. [24] addressed wells operating with Electrical Submersible Pumps (ESP), combining Response Surface Methodology with neural networks using key ESP parameters as inputs. Agwu et al. [25] employed multivariate adaptive regression splines (MARS), emphasizing interpretability and physical validation; these authors also conducted an extensive chronological review of data-driven BHP methods, systematically tabulating methodologies, dataset characteristics, and error margins.

The complexity of applying data-driven methods to real-world cases is highlighted by Antonelo et al. [2]: while BHP reproduction is possible for simulated models using only choke bean size as input, real noisy data from gas lift wells under slug flow required a much more complex framework using Echo State Networks. Firouzi and Rathnayake [26] collected surface and subsurface data from five coal seam gas wells over 18 months, demonstrating the superiority of neural networks over regression methods. Al Shehri et al. [27] and Olamigoke and Onyeali [28] both tested LSTM models alongside other methods, with the latter showing LSTM outperforming Random Forest and SVR in the Volve field. These studies demonstrate how machine learning is enabling solutions tailored to specific conditions, but they are largely limited to single-field applications with sufficient training data available.

Another group of works combines data-driven models with physical knowledge to improve accuracy and generalization. Integrating physics-based equations, computational simulations, and machine learning algorithms has demonstrated advantages [29]. Zalavadia et al. [18] developed a hybrid model that selects the most suitable physics-based correlation for each sample and then refines the estimate using physics-informed machine learning, achieving strong results with simple regression methods. Similarly, Zheng et al [6] incorporated physics-based loss functions into NN and XGBoost models, which proved important when dealing with a small dataset from two wells in a gas field at an early stage of development.

Transfer learning has recently gained traction in petroleum engineering as a strategy to overcome data scarcity in new domains. Tang et al. [30] combined LSTM and transfer learning to predict shale oil well production across fields in different countries, demonstrating that knowledge from large North American datasets improved prediction accuracy by over 50% in Chinese fields with limited history. Zhang et al. [31] addressed depth-lagged LWD gamma ray measurements by transferring models from the Volve field in Norway to a gas field in China, improving accuracy by up to 13% in data-limited environments. Zhu et al. [32] employed transfer learning with convolutional neural networks for fault parameter inversion from well testing data, showing improved accuracy and significantly reduced computation times compared to manual methods. Closer to our problem, Zhang et al. [33] proposed a hybrid LSTM-DANN transfer learning approach for BHP prediction during drilling operations, achieving MAPE of 0.15% across different well sections. However, to the best of our knowledge, the application of transfer learning to BHP estimation during production–a fundamentally different regime governed by multiphase flow dynamics and long-term reservoir depletion rather than drilling hydraulics–remains unexplored prior to the present work.

We observe, more broadly, that existing studies have not adequately addressed flowing BHP estimation under conditions of pronounced concept drift and domain shift, characterized by substantial inter-well and temporal variations in production parameters.

In our previous works, we first explored a deep learning approach using LSTM, obtaining improvements over simpler neural networks and demonstrating suitability for practical, real-time applications [34]. We then proposed an ensemble-based solution combining unsupervised clustering with supervised ML, evaluating models of different categories and complexities (neural networks, linear regression, XGBoost, and LightGBM) and showing that domain partitioning into operating regimes enables optimal performance even with simpler models [35]. A Physics-Guided Neural Network (PGNN) framework, integrating first-principles flow modeling with data-driven learning, was subsequently applied to BHP estimation in injector wells operating with variable $CO_2$ content, relevant to EOR and CCUS projects [36].

In this study, we expand the scope by incorporating a significantly broader dataset and a wider range of operational conditions. Notably, we are, to the best of our knowledge, among the first to apply transfer learning specifically to BHP estimation during production in offshore oil wells, with the goal of developing a generalized and scalable framework that can be used across multiple oilfields with varying characteristics.

## 2.4. Objectives and Contributions

This study aims to develop a methodology for estimating the flowing bottom-hole pressure in oil wells during regular operation, leveraging a data-driven approach based on supervised learning (including deep learning).

To achieve this, we implement and evaluate multiple configurations of Long Short-Term Memory (LSTM) and Multi-Layer Perceptron (MLP) neural networks, as well as regularized linear regression. Using these supervised learning methods, we intend to extract knowledge and "learn" the fluid flow dynamics in well tubing and production flowlines to estimate BHP from surface measurements and wellhead gauge data, without relying on analytical models, empirical equations, or the properties of produced fluids. This selection of methods ranges from widely-used basic techniques to a prominent method for sequential data in deep learning.

Additionally, we evaluate the benefits of transfer learning for this task, using the models trained with the dataset from an oilfield that has longer history and larger number of wells as a base model to be adapted to another field where the available data for training is limited and the production conditions are different, including the use of artificial lift.

Our motivation is to obtain a virtual meter capable of estimating BHP reliably enough to be used in well and reservoir monitoring actions, including real-time possibilities, being able to be deployed when physical sensors for bottom hole measurements are unavailable. We highlight that the proposed methodology is intended to be widely applicable across different fields and operational scenarios where similar surface measurements exist, particularly in cases where PDGs are impractical due to cost or technical constraints.

The main contributions of this work are:

- Soft sensor validated at industrial scale: we present a data-driven estimator trained exclusively on wellhead and topside measurements, validated across 45 wells spanning diverse production conditions, one of the most comprehensive real-world evaluations in the BHP estimation literature.

- Application of Transfer Learning to BHP estimation in oil wells: a fine-tuning framework that adapts a base model trained on a large source field to a target field with different reservoir and operational characteristics, without complete retraining. This extends our previous work on ensemble methods [35] and physics-guided approaches [36] by addressing the cross-field domain adaptation challenge.

- Practical deployment: millisecond inference times enable integration with real-time production monitoring systems, supporting reservoir management, flow assurance monitoring, and early anomaly detection in wells without available PDGs.

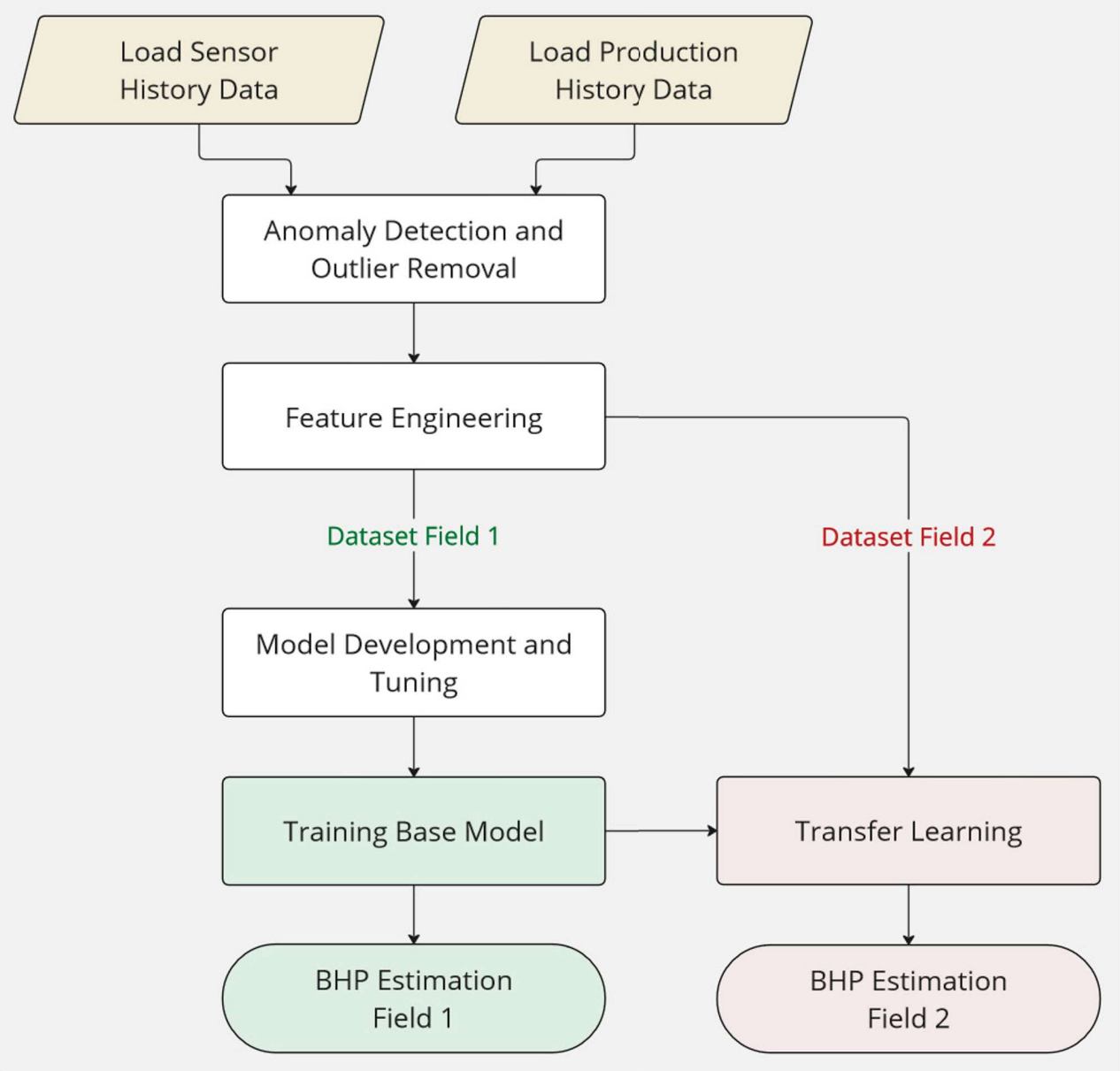

**Figure 3:** Flowchart depicting the proposed methodology, including base model training and transfer learning stages.

## 3. Methodology

This section details the data acquisition process, analyzes key attributes of the dataset, and presents the machine learning framework developed for BHP estimation. Our methodology for addressing this challenge is systematically organized in the workflow diagram provided in Figure 3.

We implemented this solution in Python, leveraging *TensorFlow*, *scikit-learn*, *XGBoost*, and *LightGBM*, along with interfaces to the Plant Information (PI) real-time monitoring system and the operator's production databases.

### 3.1. Database

We used as a case study data collected over 13 years from two distinct offshore oilfields located in the Brazilian pre-salt basin, including 9 platforms and a total of 60 oil producing wells, with 45 of them in our first oilfield, that for confidentiality reasons we will call Field 1, and the other 15 wells in the so-called Field 2.

An important characteristic of this particular dataset is the diversity of reservoir and fluid conditions across both spatial and temporal domains. This complexity is a consequence of a superposition of factors: (1) regional variations in initial fluid composition due to the reservoir's large dimensions; (2) uneven evolution of drainage progression across its different regions; (3) lithological heterogeneities; and (4) recovery strategies used over time, including Water-Alternating-Gas injection. These factors, plus differences in perforation depths, well geometries and flowline lengths, lead to a large variety of flow conditions. Such diversity results in a challenging problem, as accurate BHP estimation must hold across this plethora of conditions.

In those fields, as a project assumption all wells have one or more PDGs installed at completion, but we observed that a relevant number of these became unavailable since then. As we stated in our previous work [34], downhole gauge failure is approximately three times more frequent than its counterparts located at the wellhead, due to the harsh operating conditions (high pressures and temperatures) [6] and the longer distances for signal traveling. Also, their replacement is far more complex and expensive.

The variables we assumed as relevant to estimate BHP and that are readily available in the operator company's databases are shown in Table 1, including node of measurement and units.

Production rates are calculated daily by allocating platform-level totals to individual wells based on their latest production test results, a methodology that introduces inherent uncertainty due to the proration process. All other well parameters, including the BHP itself, are sourced from the PI monitoring system, which provides continuous, high-frequency recordings. While bottom-hole temperature (BHT) data are available, we deliberately excluded it from our analysis. This decision aligns with our objective to estimate BHP under conditions where PDG signals are absent, as the predominant failure mode [5] involves concurrent loss of both pressure and temperature measurements.

**Table 1**
List of variables used for the supervised learning.

| Variable | Node | Unit |
|---|---|---|
| Bottom-hole Pressure (BHP) | Bottom-hole (PDG) | $kgf/cm^2$ |
| Bottom-hole Temperature (BHT) | Bottom-hole (PDG) | $^oC$ |
| Wellhead Pressure (WHP) | Wellhead (TPT) | $kgf/cm^2$ |
| Wellhead Temperature (WHT) | Wellhead (TPT) | $^oC$ |
| Choke inlet pressure | Platform | $kgf/cm^2$ |
| Choke inlet temperature | Platform | $^oC$ |
| Choke valve aperture | Platform | $\%$ |
| Oil production rate | Platform | $m^3/d$ |
| Gas production rate | Platform | $Mm^3/d$ |
| Water production rate | Platform | $m^3/d$ |
| Gas lift injection rate | Platform | $Mm^3/d$ |

The two oilfields that compose our dataset present distinct characteristics in terms of location, reservoir properties, and production parameters, as we show in Table 2 with a statistical summary of the variables. A comparison of the most relevant variables for both fields is shown in the boxplots in Figure 4. We observe that, on average, Field 2 has higher pressures, gas-oil ratios (GORs), water cuts, and well depths compared to Field 1. Another important difference is that Field 1 shows minimal and almost negligible use of gas lift, while Field 2 has had wells using this method for longer periods. These characteristics provide favorable conditions for the application of a transfer learning framework, as they allow for the exploration of how data from one field can benefit the learning process for another without having to retrain an entire model, and also taking advantage of its ability to generalize across different data distributions. Choke openings greater than 100% indicate apertures exceeding the 1 inch standard, found in some wells.

As a quantitative metric to characterize the domain shift between both fields, we computed the symmetric Kullback-Leibler (KL) divergence between the empirical distributions of each variable [37]. The KL divergence measures how much one probability distribution differs from another; the symmetric version averages both directions, providing a balanced estimate:

$$KL_{sym}(P|Q) = \frac{1}{2}\sum_{x} p(x)\ln\frac{p(x)}{q(x)} + \frac{1}{2}\sum_{x} q(x)\ln\frac{q(x)}{p(x)} \quad (2)$$

where $p(x)$ and $q(x)$ are the probability density functions of Field 1 and Field 2, respectively, estimated via kernel density estimation (KDE) to avoid assumptions of normality. The results are shown in Table 3.

The largest divergences are observed for GOR and well depth, reflecting the distinct reservoir maturity and completion characteristics of the two fields. Gas lift rate, water cut, and bottom-hole pressure itself also show substantial domain shift, confirming that Field 2 operates under a fundamentally different production regime. Oil production rate, by contrast, presents near-identical distributions, providing a common operational basis that facilitates knowledge transfer.

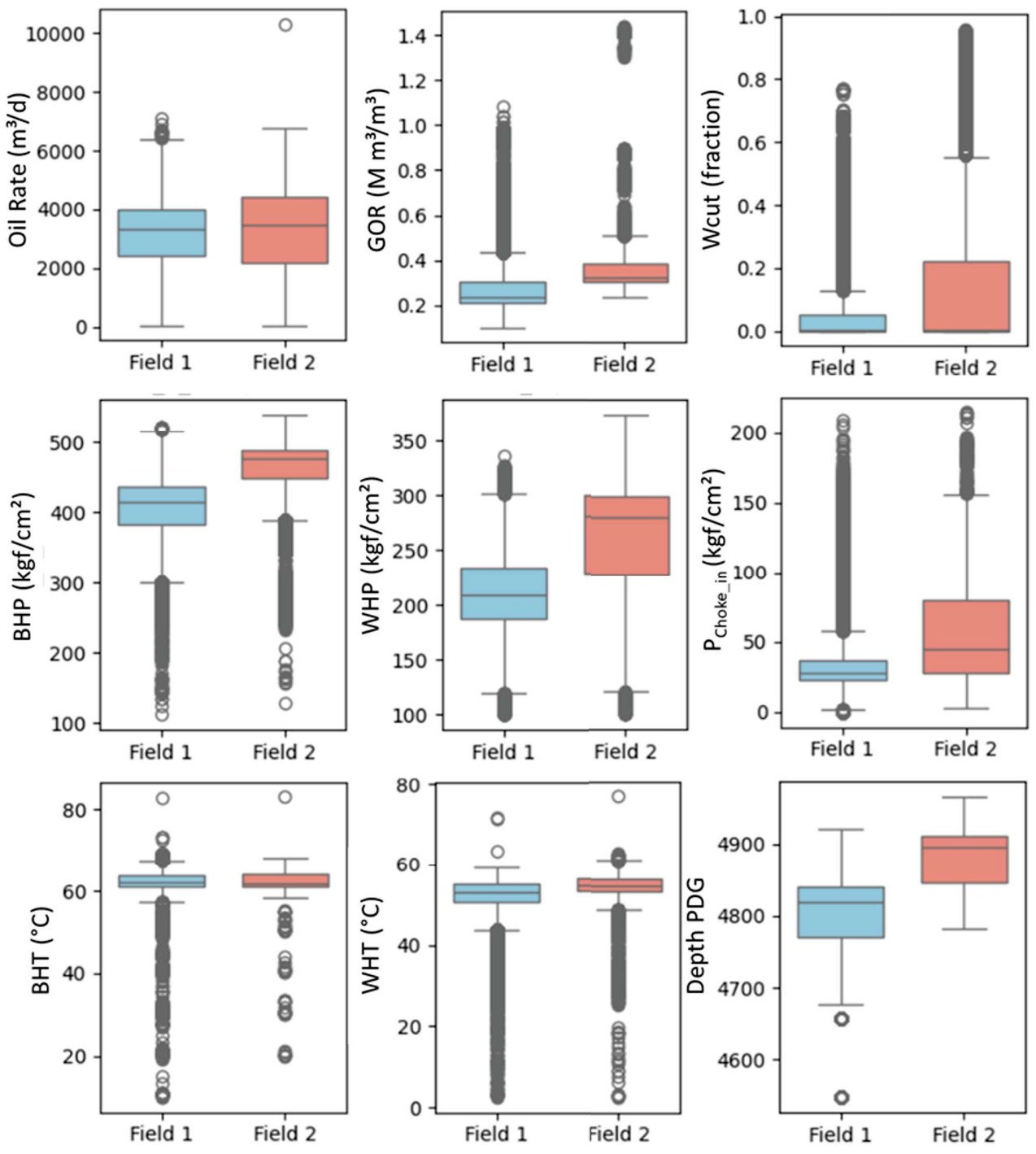

**Figure 4:** Boxplots showing the probability distributions of the variables for the two fields.

We highlight that our dataset encompasses a wide array of operational conditions and production parameters for both fields, being one of the largest (with more than 100,000 samples) and most comprehensive in the related literature, as we can see when compared to diverse compiled references [25]. Those characteristics are particularly important when dealing with deep learning methods that are data-intensive. This dataset offers a solid foundation for validating the methodology and demonstrates its potential for application to other platforms or fields.

**Table 2**
Statistical summary of the dataset.

| Stats. | Field | Depth PDG ($m$) | Oil rate ($m^3/d$) | Gas rate ($Mm^3/d$) | Water rate ($m^3/d$) | Choke Open. (%) |
|---|---|---|---|---|---|---|
| mean | 1 | 4804.7 | 3168.9 | 832.3 | 150.3 | 68.3 |
| mean | 2 | 4879.4 | 3169.2 | 1075.9 | 532.6 | 48.8 |
| std-dv | 1 | 65.7 | 1134.6 | 296 | 307.5 | 37.1 |
| std-dv | 2 | 49.8 | 1483.9 | 504.4 | 922.2 | 40.5 |
| min | 1 | 4547.5 | 52.6 | 9.6 | 0 | 4 |
| min | 2 | 4782.3 | 55.6 | 30.2 | 0 | 4 |
| max | 1 | 4921.7 | 7111.8 | 2150.2 | 2776.9 | 125 |
| max | 2 | 4966.3 | 10294.3 | 3284.1 | 5622.3 | 103.2 |
| **Stats.** | **Field** | **Press. Choke in. ($kgf/cm^2$)** | **Temp. Choke in. ($^oC$)** | **WHP ($kgf/cm^2$)** | **WHT ($^oC$)** | **BHP ($kgf/cm^2$)** |
| mean | 1 | 38.8 | 28.1 | 207.3 | 52.5 | 404.1 |
| mean | 2 | 58.9 | 34 | 264.1 | 54.8 | 462.2 |
| std-dv | 1 | 28.9 | 8.3 | 40.2 | 4.2 | 48.8 |
| std-dv | 2 | 37.6 | 7 | 46.3 | 3.8 | 45.1 |
| min | 1 | 0 | 0.1 | 100 | 2.6 | 111.6 |
| min | 2 | 3.1 | 0.6 | 100.1 | 2.5 | 127.8 |
| max | 1 | 208.4 | 100 | 335.9 | 71.8 | 521.9 |
| max | 2 | 214.1 | 82.1 | 375.3 | 76.8 | 540.7 |

**Table 3**
Symmetric KL divergence between Field 1 and Field 2 distributions, estimated via kernel density estimation.

| Variable | $KL_{sym}$ |
|---|---|
| GOR | 3.160 |
| Depth PDG | 2.837 |
| Gas lift injection rate | 1.466 |
| Water cut | 1.427 |
| Bottom-hole Pressure (BHP) | 1.298 |
| Wellhead Pressure (WHP) | 1.160 |
| Wellhead Temperature (WHT) | 0.854 |
| Choke inlet temperature | 0.484 |
| Choke inlet pressure | 0.315 |
| Oil production rate | 0.170 |

## 3.2. Data Preprocessing and Feature Engineering

Once the scope of our problem is defined, we start the sequence of steps for data preprocessing and feature engineering, fundamental to feed the ML methods with quality data, maximizing their potential.

1) Our first step is loading the datasets using a Python API that integrates with the company's databases for production records and with the PI monitoring system. The latter is capable of providing real-time monitoring of the most relevant sensors in the production system, registering every second–we export as hourly averages to better suit our purpose. Production data are registered with daily periodicity, where official measurement is performed for the total volume of the platform, and a share is attributed to each well based on its most recent production test. These tests are made by directing the production of a single well to a test separator to have its production rates monitored for a determined time [17]. This can be considered a drawback and a limitation for the accuracy of our estimators, since alterations in individual well behavior can remain unnoticed until the next test, which can take up to three months.

2) Data cleaning is performed by removing all samples with at least one null or error value, those with production rates or choke aperture equal to zero (to ensure we deal only with flowing BHP), and the ones where the well remained open for less than 2 hours during a day, which could represent transients.

3) We implemented a simple anomaly detector to remove "frozen" sensor data, actuating when the measured values remain erroneously constant for more than 3 days.

4) We aligned all variables to the same time base, calculating daily averages for the sensor data. This resulted in 69,039 total samples from Field 1, and 29,799 from Field 2.

5) We added to the dataset derived variables calculated from production rates, the GOR:

$$GOR = \frac{Q_g}{Q_o} \tag{3}$$

and the Water Cut:

$$W_{cut} = \frac{Q_w}{(Q_o + Q_w)} \tag{4}$$

where $Q_o$, $Q_g$, and $Q_w$ represent oil, gas, and water production rates, respectively. The new variables can be used as alternatives to gas and water rates in our dataset, with lower variability.

6) The next step is dataset partitioning. For Field 1, we separated two different subsets for blind tests: one consisting of the entire history for a group of 9 wells chosen randomly, and the other with the last year of history of all wells, simulating a future application scenario. The remaining data are used for training and validation, segmented for 5-fold cross validation [38]. Figure 5 illustrates this partitioning.

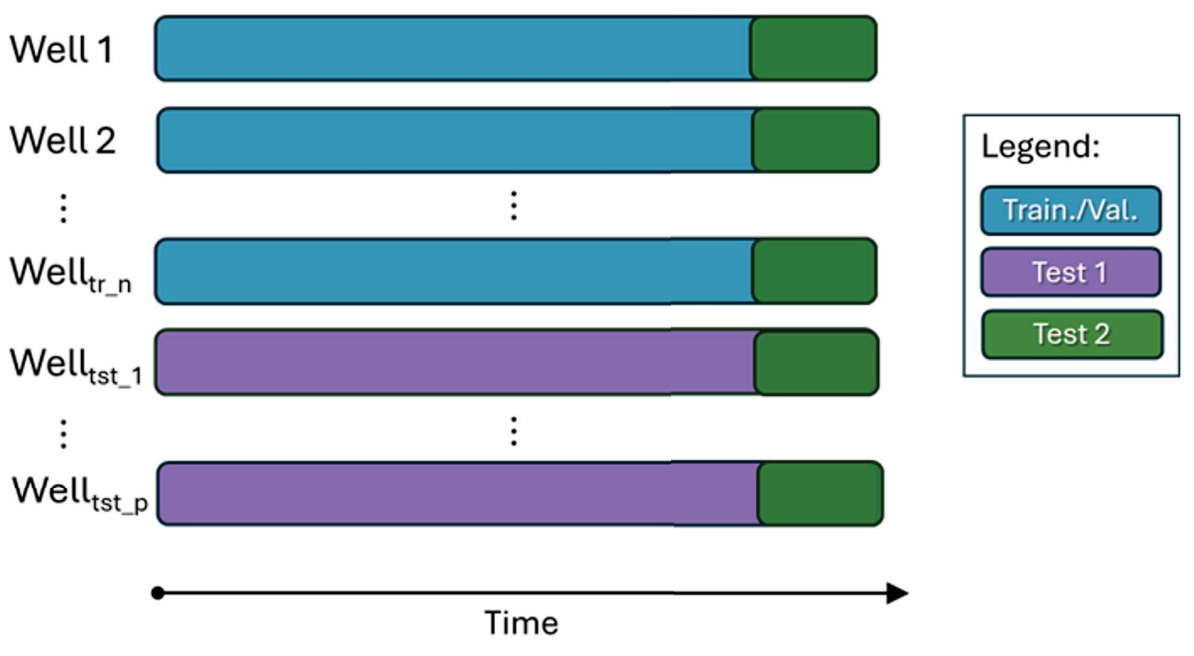

**Figure 5:** Dataset partitioning, creating two test subsets.

For Field 2, consisting of only two platforms, we separated data from one of them for testing (7 wells), and also used the last year of all wells as a second test dataset. The remaining data are used for training, considering that despite the limited number of samples, we can benefit from the transfer learning framework using knowledge from Field 1.

7) Outlier removal is based on Interquartile Ranges (IQR) to remove a sample $i$ if any of its variables $x_i$ satisfies the conditions for the lower limit:

$$x_i < Q_1 - 1.5 \cdot IQR \quad (5)$$

and the higher limit:

$$x_i > Q_3 + 1.5 \cdot IQR \quad (6)$$

where $Q_1$ and $Q_3$ are the values for which the Cumulative Distribution Function for $x$ equals to 25% and 75%, respectively, and $IQR = Q_3 - Q_1$. Exceptions are made if deviations slightly larger than the threshold are physically plausible, e.g., oil rates in wells operating temporarily with restrictions. We perform that step separately for our two datasets, given that the fields have different statistical distributions.

8) An analysis of the correlations among the variables can be used to support the selection of the best set of inputs for our regression problem. Using the training/validation dataset for Field 1, we generated the correlation heatmap shown in Figure 6, consisting of values calculated using Pearson's coefficient [39]:

$$r = \frac{\sum_{i=1}^{n}(x_i - \bar{x}) \cdot (y_i - \bar{y})}{\sqrt{\sum_{i=1}^{n}(x_i - \bar{x})^2 \cdot \sum_{i=1}^{n}(y_i - \bar{y})^2}} \quad (7)$$

where $r$ is the correlation coefficient, $x_i$ and $y_i$ the samples of two different sets of variables, $\bar{x}$ and $\bar{y}$ their respective mean values, and $n$ the number of samples.

9) Next, we performed data normalization, evaluating two different approaches: Gaussian normalization using mean and standard deviation, and the *MinMax* normalization limiting the ranges of all variables to the interval [0, 1].

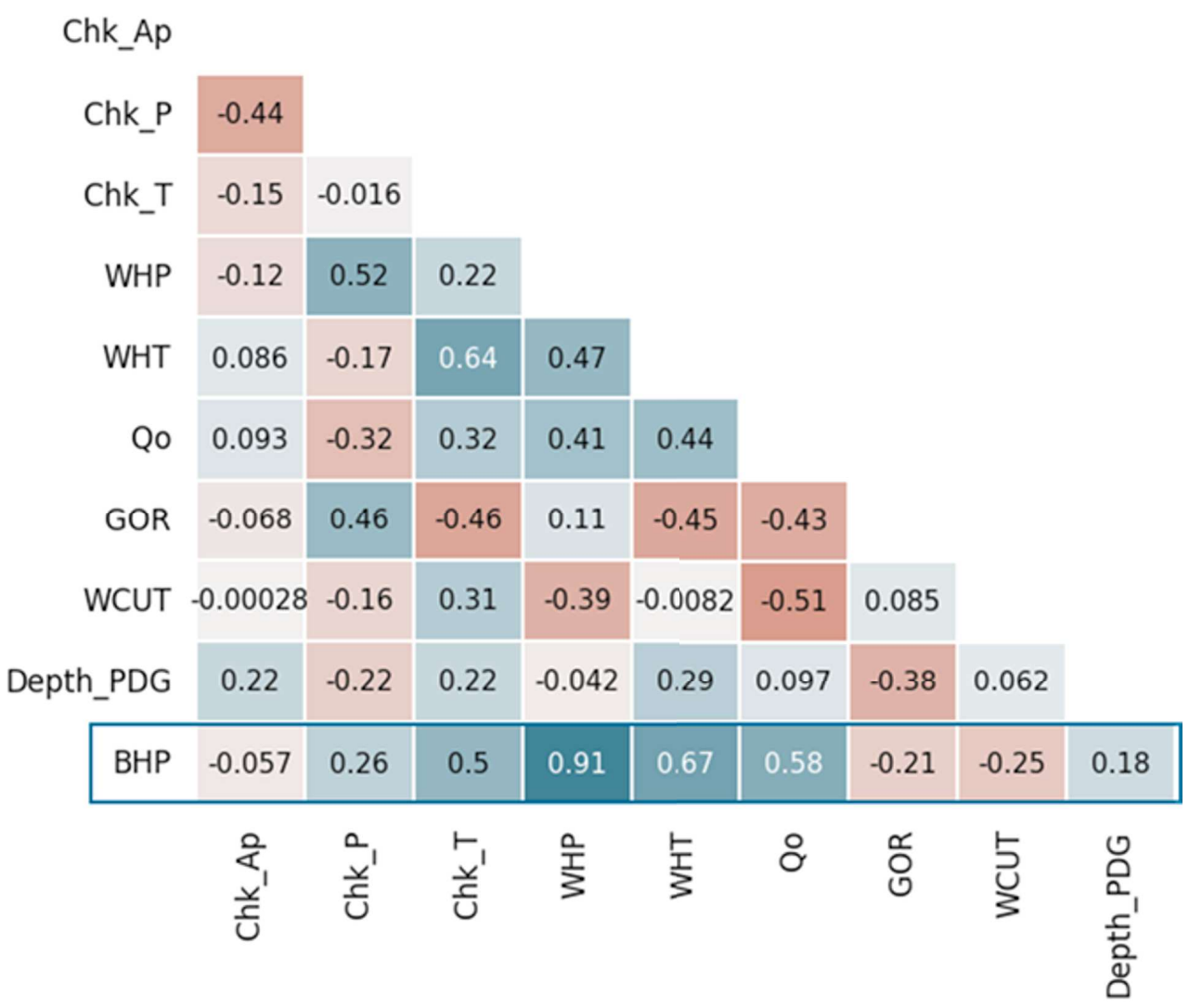

**Figure 6:** Correlation heatmap for the variables of the problem.

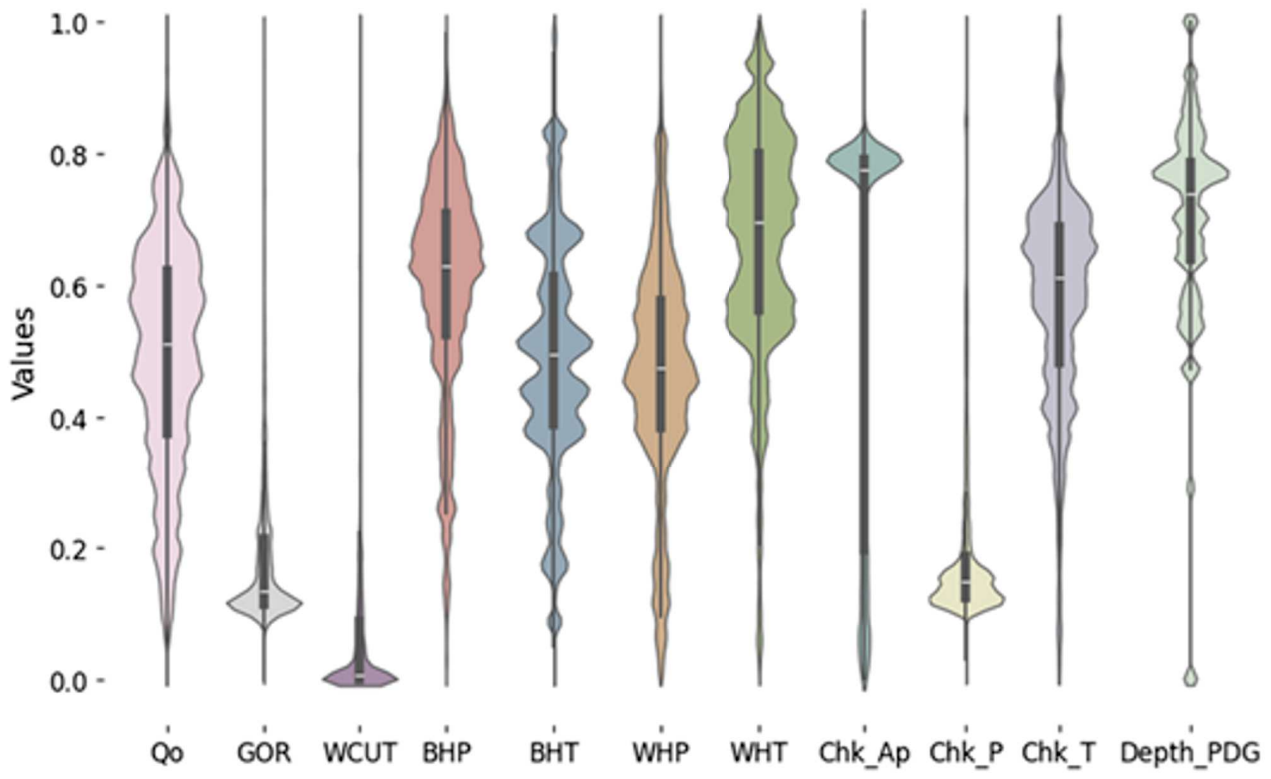

**Figure 7:** Violin plots of the variables after *MinMax* normalization for Field 1.

The latter can avoid extreme value ranges compared to the first approach, but at the same time it can compress relevant values to a narrower range [40]. Normalization parameters are calculated using only training data (separately for the two datasets), preventing information leakage from the validation and test sets. Violin plots showing the distribution of the variables after this step, for the example of Field 1 with *MinMax* normalization, are presented in Figure 7.

10) Finally, we construct time-dependent input samples for the LSTM model. For this, each sample will carry values for a number of previous time steps, as we exemplify in Figure 8. There, we have $n$ input variables and $p$ previous time steps feeding the LSTM to estimate BHP at a time $t$. For the other benchmark methods, not time-related, we consider $p = 0$ and have as inputs only the variables measured at the current time.

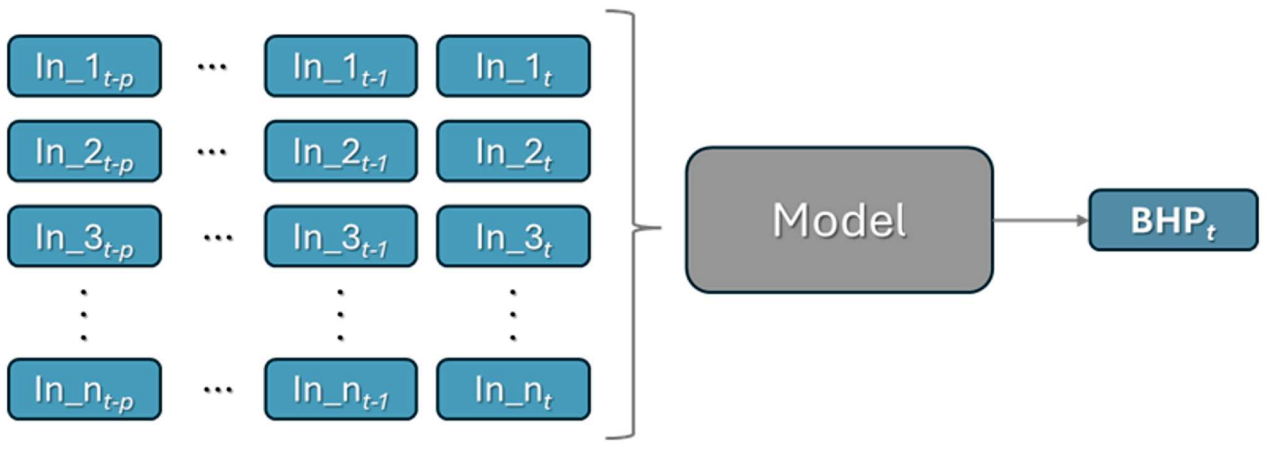

**Figure 8:** Input samples preparation for the LSTM model, illustrating the use of previous time steps to estimate the current BHP.

## 3.3. Aspects of the Machine Learning Methods

In this work, we evaluated three machine learning methods to address the proposed task. A concise overview of those methods is provided, with comprehensive implementation details available in the referenced literature. We also bring some considerations about transfer learning and its applications.

1) *Multivariable Linear Regression*: This method serves as our first benchmark due to its simplicity, low computational cost, and interpretability. Linear regression assumes that the outputs can be expressed as a weighted sum of the input features [41]. Model training consists of determining optimal weights that minimize a loss function applied to the training dataset [42].

Variants of linear regression can include regularization terms that act as penalties for the regression coefficients. This approach helps prevent overfitting by reducing the influence of input variables that may capture noise or outliers, promoting models with improved generalizability. Regularization thereby balances predictive accuracy with computational tractability, valuable in scenarios with limited training data or high-dimensional input spaces.

Ridge regression adds a term in the equation of the loss function (that will be minimized in the training phase), as shown below:

$$f_{ridge} = \sum_{i=1}^{n}[y_i - (w \cdot x_i + b)]^2 + \alpha \cdot \sum_{j=1}^{p} w_j^2 \tag{8}$$

where $y$ is the predicted value, $x$ is the input value, $w$ and $b$ the coefficients to be adjusted. The last term, responsible for the regularization, is weighted by a factor $\alpha$, which can be tuned to control the penalties of the regression coefficients. Setting $\alpha = 0$ reduces the equation to the simple linear regression, while large values may exaggeratedly shrink the coefficients [43]. We use the regression based on Equation (8), with $\alpha$ as a hyperparameter adjusted during the training and validation of the model.

2) *Neural Networks*: As our second benchmark model, we selected a NN using the traditional MLP configuration. This form of NN is one of the most popular machine learning methods due to its ability to solve non-linear problems, and helped set the foundations on which modern methods and algorithms lay. Drawing inspiration from biological nervous systems, artificial neural networks emulate synaptic connectivity using computational units called artificial neurons. Each neuron processes input signals through weighted aggregation, followed by non-linear transformation via activation functions to generate outputs [42].

By interconnecting these neurons, or nodes, in layered architectures, diverse network topologies emerge with distinct computational capabilities. The learning process involves iterative optimization of connection weights through backpropagation algorithms, enabling the network to reproduce input-output relationships present in training data.

The MLP model we adopted consists of a feed-forward structure with neurons arranged in sequential layers, trained using the backpropagation method.

3) *Long Short-Term Memory*: As our most advanced model, we selected the LSTM network due to its superior ability to capture temporal patterns in sequential data.

The Long Short-Term Memory networks belong to the family of recurrent neural networks (RNNs), which are specifically designed to retain information across time steps, making them especially effective for sequential data such as time series [41]. LSTMs were developed as a solution to the vanishing gradient problem that limits the performance of standard RNNs [44]. Their architecture incorporates both long-term and short-term memory units, as well as a set of gating mechanisms (input, forget, and output gates) that regulate the flow of information. These gates determine when to store, discard, or retrieve information from memory, allowing the model to capture dependencies over sequences [41, 44].

Although the objective of this work is to estimate bottom-hole pressure at the same time step as the input variables, we believe that the use of a sequential architecture can provide greater stability and robustness to the predictions. By capturing temporal dependencies, the model can identify dynamic patterns that might not be apparent in static approaches, resulting in more consistent and reliable predictions. The key hyperparameter for capturing this temporal dependency is the number of time steps used for the input variables, which distinguishes this approach from conventional neural networks.

4) *Transfer Learning*: This relatively recent paradigm in artificial intelligence is based on the idea of using knowledge acquired from one dataset or domain as a foundation or starting point to accelerate and improve training on a different but related problem [45, 46]. This approach is especially useful when working with deep learning models, which typically require large amounts of training data to perform effectively. The framework involves training a machine learning model on a large, often more general dataset, then freezing its parameters (e.g., connection weights) and subsequently adapting the model to a new problem—one that usually lacks sufficient data for traditional deep learning methods, or when data can be easily outdated [45]. The adaptation step in supervised learning problems can be carried out through different approaches, as briefly outlined below [47]:

- Fine-tuning: This involves "unfreezing" some or all layers of the base model and retraining it using the new

dataset, typically with fewer epochs and/or a smaller learning rate than the initial training phase.

- Addition of new layers: In this approach, the base model remains unchanged, and one or more new hidden layers are added. Only the weights of these new layers are adjusted during the additional training.

Both approaches aim to preserve the general features learned by the base model while enabling efficient adaptation to the new problem. Further discussion regarding categories and different approaches to transfer learning can be found in the literature [45, 46].

Transfer learning has gained popularity in both scientific and industrial applications, particularly in tasks involving computer vision, natural language processing, and time series modeling [47]. For this latter class of problems, integration with LSTM architecture has also been actively explored in the literature; application examples range from financial market forecasting [48] to energy demand prediction [49].

In this work, we explore transfer learning for domain adaptation of our BHP estimators, using base models trained on data from a field with a larger number of samples and greater variability. We then test different adaptation strategies for applying these models to a new field with slightly different characteristics. These steps are illustrated in the final blocks of the flowchart in Figure 3.

### 3.4. Evaluation Metrics

The performance of the estimators was quantitatively evaluated using the metrics Mean Absolute Percentage Error (MAPE), Symmetric Mean Absolute Percentage Error (SMAPE), and Root Mean Squared Error (RMSE), as defined respectively by the equations below:

$$MAPE = \frac{1}{n} \cdot \sum_{j=1}^{n} \left| \frac{y_j - \hat{y}_j}{y_j} \right| \tag{9}$$

$$SMAPE = \frac{1}{n} \cdot \sum_{j=1}^{n} \frac{|y_j - \hat{y}_j|}{\frac{(y_j + \hat{y}_j)}{2}} \tag{10}$$

$$RMSE = \sqrt{\frac{1}{n} \cdot \sum_{j=1}^{n} (y_j - \hat{y}_j)^2} \tag{11}$$

where $y_j$ are the predicted values and $\hat{y}_j$ the actual values of the dependent variable [50], in our case, the bottom-hole pressure. For consistency with the other metrics and to allow comparison across datasets with different pressure ranges, we report the normalized RMSE (nRMSE), obtained by dividing the RMSE by the mean of the actual values.

Qualitatively, we will also examine the curves of actual and estimated pressure values over time, assessing the estimators' ability to capture and reproduce trends in pressure behavior, such as during periods of well restrictions, production decline, or response to injection effects.

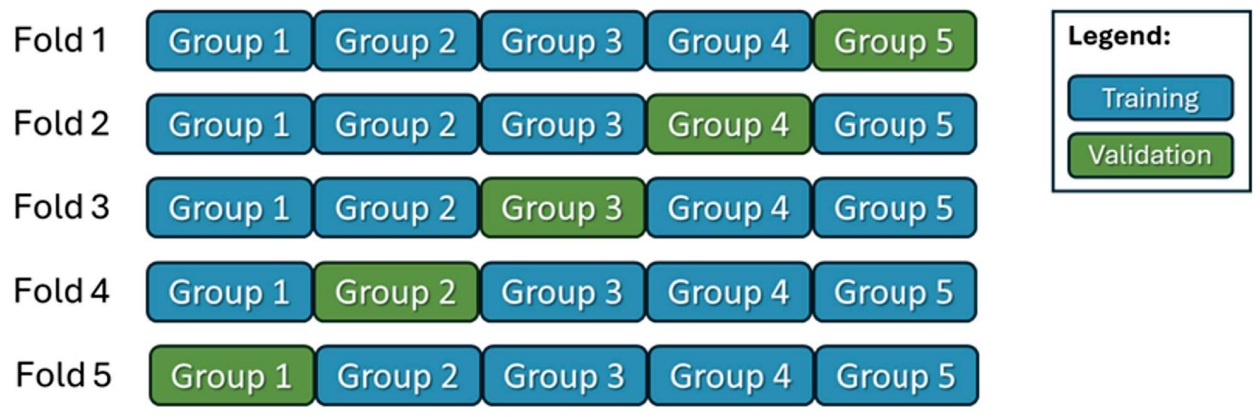

**Figure 9:** Dataset preparation for training and 5-fold cross-validation.

## 4. Results

We present the results in three phases: in the first we tune hyperparameters for the LSTM and the benchmark NN and ridge regression models using training and validation subsets from our larger oilfield; in the second we evaluate performance for a test subset from the same field, and in the third we evaluate transfer learning for our new oilfield. Both tests simulate real-life situations for oil wells without available PDG data.

### 4.1. Hyperparameter Tuning and Model Selection

For hyperparameter tuning and defining the optimal configurations of each selected model, we relied on the training and validation subset extracted from Field 1 data, as illustrated earlier in Figure 5. We employed a 5-fold cross-validation scheme, where wells were randomly assigned to each fold and remained unchanged during model tuning and validation. In each step, estimators were trained on four folds and validated on the fifth, as depicted in Figure 9. Performance metrics were computed using Equations (9), (10), and (11), and final metrics were obtained by averaging results across all folds.

To account for the stochastic nature of ML model training, each configuration was run multiple times, and standard deviations were recorded for subsequent analysis.

This procedure was repeated for every tested model configuration to identify optimized hyperparameter combinations. The search was conducted through systematic grid exploration within the ranges specified in Tables 4 and 5, guided by domain experience to prioritize configurations with adequate complexity for the problem.

For the linear regression model, only the regularization coefficient $\alpha$ was tunable. Neural networks required optimization of network architecture (e.g., layers, nodes), activation functions, loss functions, and learning rate. For LSTM, we added the number of time steps per sample as a tunable parameter. For both NN and LSTM, we evaluated the inclusion of dropout—a regularization technique where randomly selected nodes are temporarily deactivated (weights set to zero before calculating the subsequent layer) during training to mitigate overfitting [41].

Tables 4 and 5 summarize the adjustable parameters and value ranges tested during optimization for the NN and LSTM models, respectively.

**Table 4**
Hyperparameters and their respective ranges evaluated in our neural network models.

| Parameter | Values (ranges) |
|---|---|
| Number of layers | 1 to 3 |
| Number of neurons | 20 to 300 |
| Dropout | 0.05 to 0.50 |
| Loss function | MSE, MAE, SMAPE |
| Activation function | ReLU, GELU, ELU, Tanh |
| Learning rate | 0.0001 to 0.003 |

**Table 5**
Hyperparameters and their respective ranges evaluated in our LSTM models.

| Parameter | Values (ranges) |
|---|---|
| Number of layers | 1 to 3 |
| Number of elements | 20 to 200 |
| Dropout | 0.05 to 0.50 |
| Error function | MSE, MAE, SMAPE |
| Activation function | ReLU, Tanh |
| Timesteps | 2 to 15 |

**Table 6**
Sets of input variables tested in the model definition.

| Set # | Input variables |
|---|---|
| Set 1 | [Choke P, Choke T, WHP, WHT, $Q_o$, $Q_g$, Depth PDG] |
| Set 2 | [Choke P, Choke T, WHP, WHT, $Q_o$, $Q_g$, $Q_w$, Depth PDG] |
| Set 3 | [Choke P, Choke T, WHP, WHT, $Q_o$, GOR, $W_{cut}$, Depth PDG] |
| Set 4 | [Choke Ap, Choke P, Choke T, WHP, WHT, $Q_o$, $Q_g$, Depth PDG] |
| Set 5 | [Choke P, WHP, $Q_o$, GOR, $W_{cut}$, Depth PDG] |
| Set 6 | [Choke P, WHP, $Q_o$, $Q_g$, $W_{cut}$, Depth PDG] |

Finally, we assessed different input variable combinations, including the full feature set but also testing subsets that excluded variables with possible lower relevance, based on the correlation analysis in Figure 6 and on our experience. The proposed configurations are displayed in Table 6. These configurations constitute an implicit ablation study of feature relevance, systematically evaluating the contribution of different variable groups to model performance.

To illustrate the results of the search for optimal configurations, Table 7 presents the mean (after multiple runs) of the three error metrics for the top-performing models in our validation set, whose hyperparameters are described below:

- **LR1**: Set 3, *MinMax* scaler, $\alpha = 0.2$.
- **NN1**: Set 5, *Standard* scaler, 1 layer, 60 neurons, 0.26 dropout, loss MSE, activation *ReLU*.
- **NN2**: Set 3, *Standard* scaler, 1 layer, 100 neurons, 0.15 dropout, loss MSE, activation *ReLU*.

**Table 7**
Results obtained with the most representative models for the 5-fold validation.

| Model | MAPE | SMAPE | nRMSE |
|---|---|---|---|
| LR1 | 1.363 | 1.359 | 2.047 |
| NN1 | 1.437 | 1.432 | 2.07 |
| NN2 | 1.466 | 1.466 | 2.1 |
| LSTM1 | **1.349** | **1.347** | **1.967** |
| LSTM2 | 1.398 | 1.395 | 2.013 |
| LSTM3 | 1.501 | 1.503 | 2.108 |

- **LSTM1**: Set 3, *MinMax* scaler, 2 timesteps, 1 layer, 80 elements, 0.30 dropout, loss MSE, activation *Tanh*.
- **LSTM2**: Set 3, *MinMax* scaler, 3 timesteps, 1 layer, 40 elements, 0.30 dropout, loss MSE, activation *Tanh*.
- **LSTM3**: Set 3, *MinMax* scaler, 10 timesteps, 1 layer, 40 elements, 0.30 dropout, loss MSE, activation *Tanh*.

Key Observations:

- Performance metrics were very similar across all methods and configurations.
- Linear Regression: Demonstrated minimal sensitivity to the $\alpha$ parameter, likely due to the dataset characteristics (large sample size, relatively few outliers, and low dimensionality).
- Neural Networks: Models with a single hidden layer outperformed deeper architectures, showing less tendency toward overfitting.
- LSTM: Achieved the best performance, although by a very small margin over the simpler methods.
- The three LSTM configurations also constitute a time-step sensitivity analysis, covering short (p=2), medium (p=3), and long (p=10) temporal windows. Configurations with a lower number of time steps performed better, in general. This is consistent with the daily resolution of the dataset: BHP in producing wells responds rapidly to changes in surface conditions, typically equilibrating within one to two days. Longer temporal windows introduce older observations that may reflect a different operating regime, adding noise rather than useful signal. For higher-frequency data, longer time steps might prove more beneficial.
- As for the input variables, Set 3 showed as the best for most configurations, indicating that almost all input variables can add valuable information for the estimation.
- Normalization with *MinMax* scaler was the best option for both LR and LSTM, while NN worked better with *Standard* scaler.

These six top-performing models were selected for further evaluation using the blind test dataset in the next phase.

**Table 8**
Results obtained with the most representative models on the first blind test dataset, for Field 1.

| Model | Test Set 1 (9,410 samples) | | |
|---|---|---|---|
| | MAPE | SMAPE | nRMSE |
| LR1 | **1.244 ±0.0** | **1.256 ±0.0** | **1.997 ±0.0** |
| NN1 | 1.376 ±0.106 | 1.393 ±0.108 | 2.174 ±0.094 |
| NN2 | 1.341 ±0.160 | 1.358 ±0.164 | 2.062 ±0.137 |
| LSTM1 | 1.309 ±0.071 | 1.326 ±0.073 | 2.075 ±0.055 |
| LSTM2 | 1.452 ±0.233 | 1.471 ±0.239 | 2.188 ±0.185 |
| LSTM3 | 1.468 ±0.106 | 1.491 ±0.109 | 2.248 ±0.110 |

**Table 9**
Results obtained with the most representative models on the second blind test dataset, for Field 1.

| Model | Test Set 2 (7,312 samples) | | |
|---|---|---|---|
| | MAPE | SMAPE | nRMSE |
| LR1 | 1.355 ±0.0 | 1.359 ±0.0 | 1.976 ±0.0 |
| NN1 | 1.338 ±0.072 | 1.337 ±0.076 | 1.974 ±0.058 |
| NN2 | 1.115 ±0.082 | 1.113 ±0.082 | 1.770 ±0.095 |
| LSTM1 | 1.044 ±0.031 | 1.040 ±0.032 | 1.695 ±0.025 |
| LSTM2 | 1.091 ±0.082 | 1.087 ±0.082 | 1.726 ±0.071 |
| LSTM3 | **1.022 ±0.053** | **1.018 ±0.054** | **1.626 ±0.053** |

## 4.2. Model Testing for Field 1

For testing our models on Field 1, we utilized the two reserved datasets described earlier (Figure 5): the first set comprised 9 wells with their complete historical data (excluding the last year), and the second set contained the last year of sensor and production data from all 45 wells in the original dataset. The results from the optimized models (defined in the previous section) are presented in Tables 8 and 9, respectively, for those two datasets. These results show the mean and standard deviation across multiple training runs, reflecting the stochastic nature of the models.

For the best-performing configuration of each estimator type, we plotted the estimated vs. measured value correlations, presenting them in Figures 10, 11, and 12 for the LR1, NN2, and LSTM1 models, respectively. We included on those references for the error margins, with dashed lines indicating ±10% and dash-dotted lines indicating ±5% deviations.

As key observations, we highlight stability, minimal outliers, and low error rates, especially compared to most literature benchmarks [25]. More specifically, we have:

- Good metrics were obtained with the simplest models, with most test samples exhibiting very low errors. This causes the metrics to show little variation when more complex models are introduced.
- Although with very close values, the simplest model, LR, showed the best performance for Test Dataset 1.
- For Test Dataset 2, composed of samples from a wider variety of wells and after a temporal cutoff, the LSTM model showed an advantage by a more significant margin.

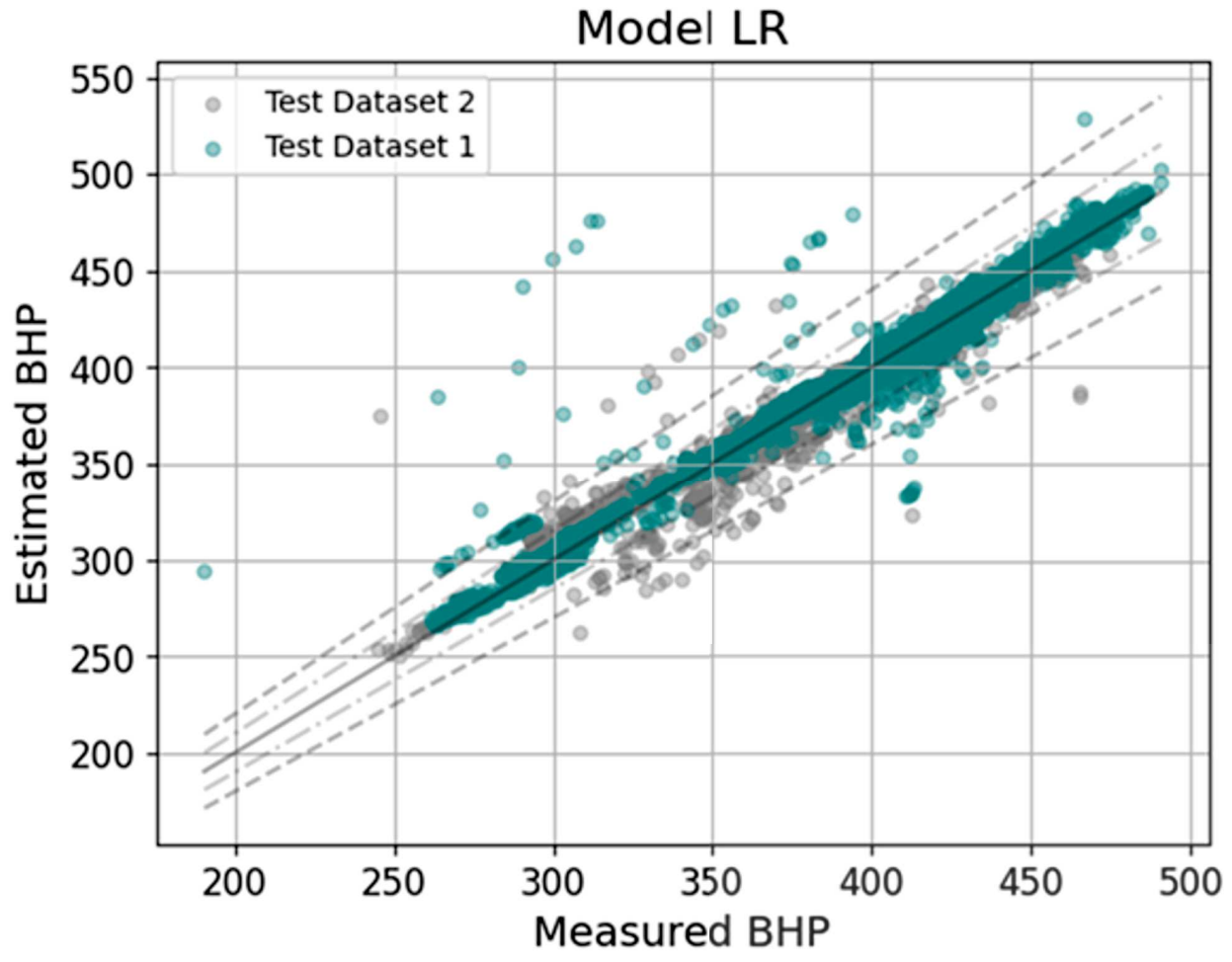

**Figure 10:** Correlation of estimated versus actual values for the BHP from our best linear regression model (Field 1). Pressure in kgf/cm².

- We can observe a slight difference in the dispersion of the estimated samples in relation to the reference line, with an advantage for the LSTM model.

Additionally, we plotted specific examples of temporal analysis for two representative wells, for which we compared measured data with estimates from the top-performing ML models from each of the families (LR, NN, and LSTM). This qualitative approach aligns closely with practical field applications, and is shown in Figure 13 and 14 for two challenging examples.

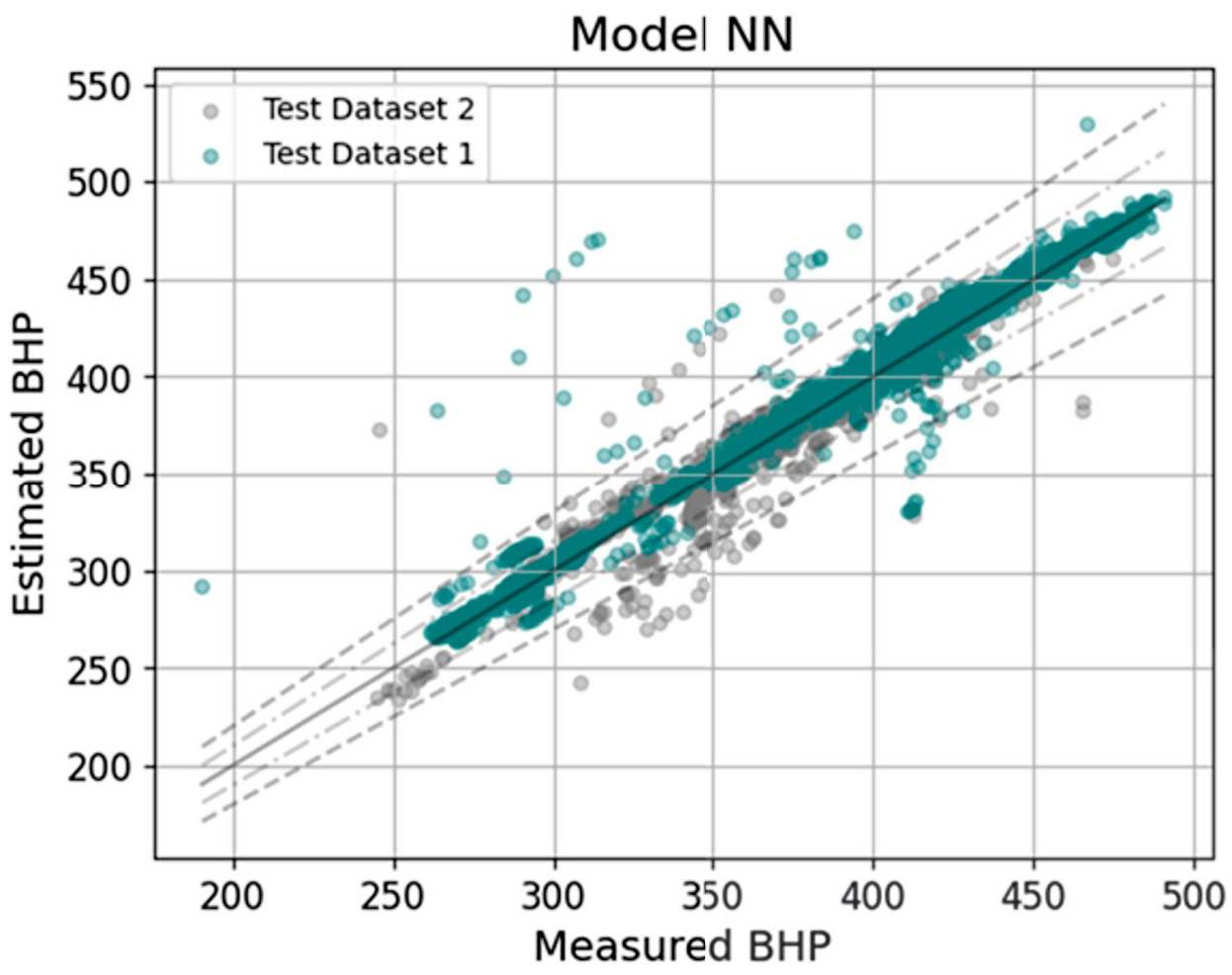

**Figure 11:** Correlation of estimated versus actual values for the BHP from our best neural network model (Field 1). Pressure in kgf/cm².

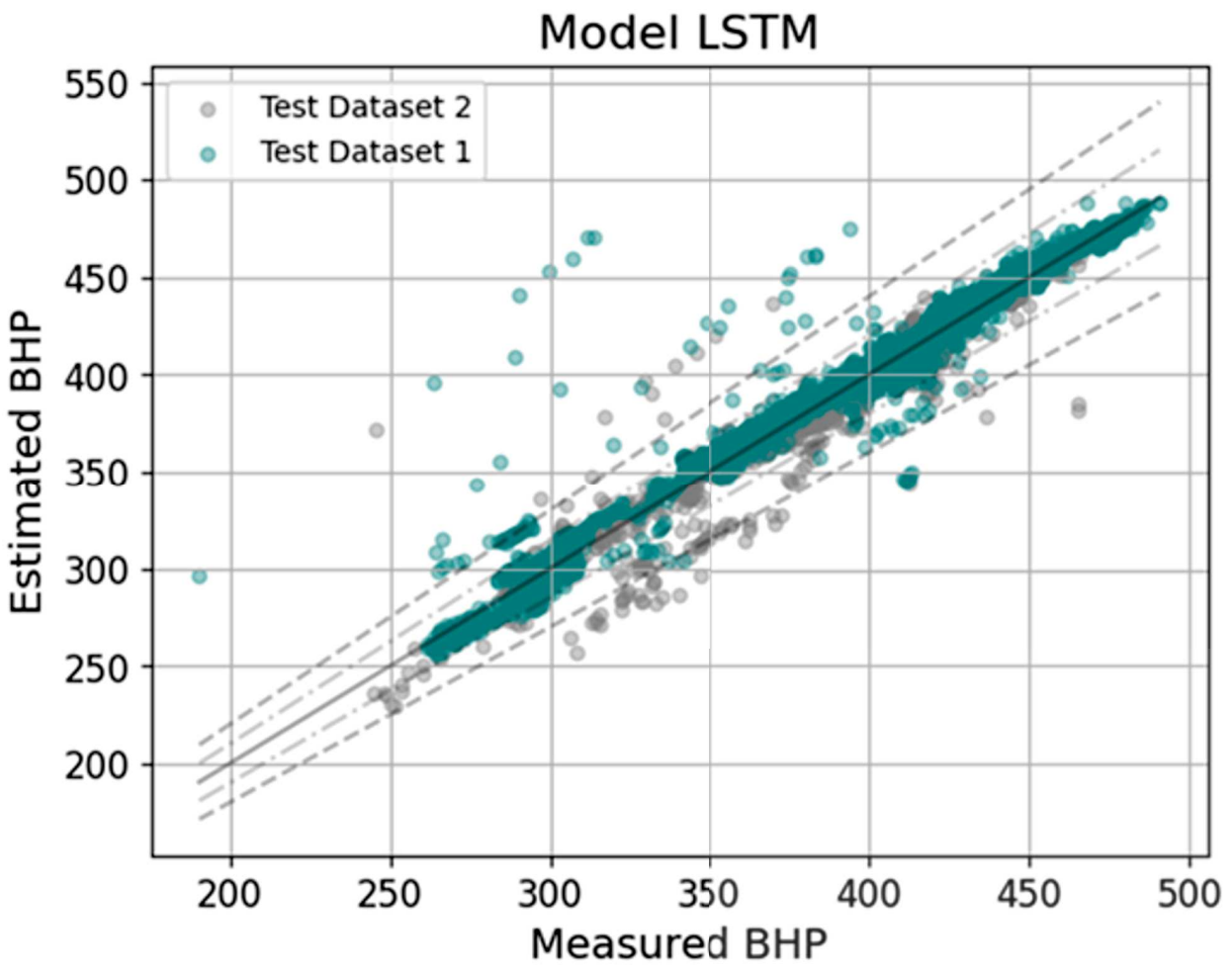

**Figure 12:** Correlation of estimated versus actual values for the BHP from our best LSTM model (Field 1). Pressure in kgf/cm².

The qualitative analysis consistently corroborates the quantitative findings, demonstrating that our top-performing models accurately capture pressure behavior, including transient events like production peaks and operational changes, while maintaining temporal stability. These results are particularly encouraging given the inherent measurement uncertainties present in field data.

One important aspect to be noted is that the LSTM model performs better than the others when we're dealing with extreme conditions, as we can see in our examples. The first (Figure 13) consists of a well with a long production history and abrupt changes in production behavior due to Inflow Control Valve (ICV) adjustments, while the second well (Figure 14) underwent extreme modulation using the choke valve throughout the analyzed period, in addition to experiencing an increasing GOR. This qualitative observation supports the use of the more complex model in scenarios where rapid operational changes are expected, even when average metrics across the full test set show modest differences. These are also relevant examples of concept drift, where the gradual change in production parameters alters the pressure dynamics over time.

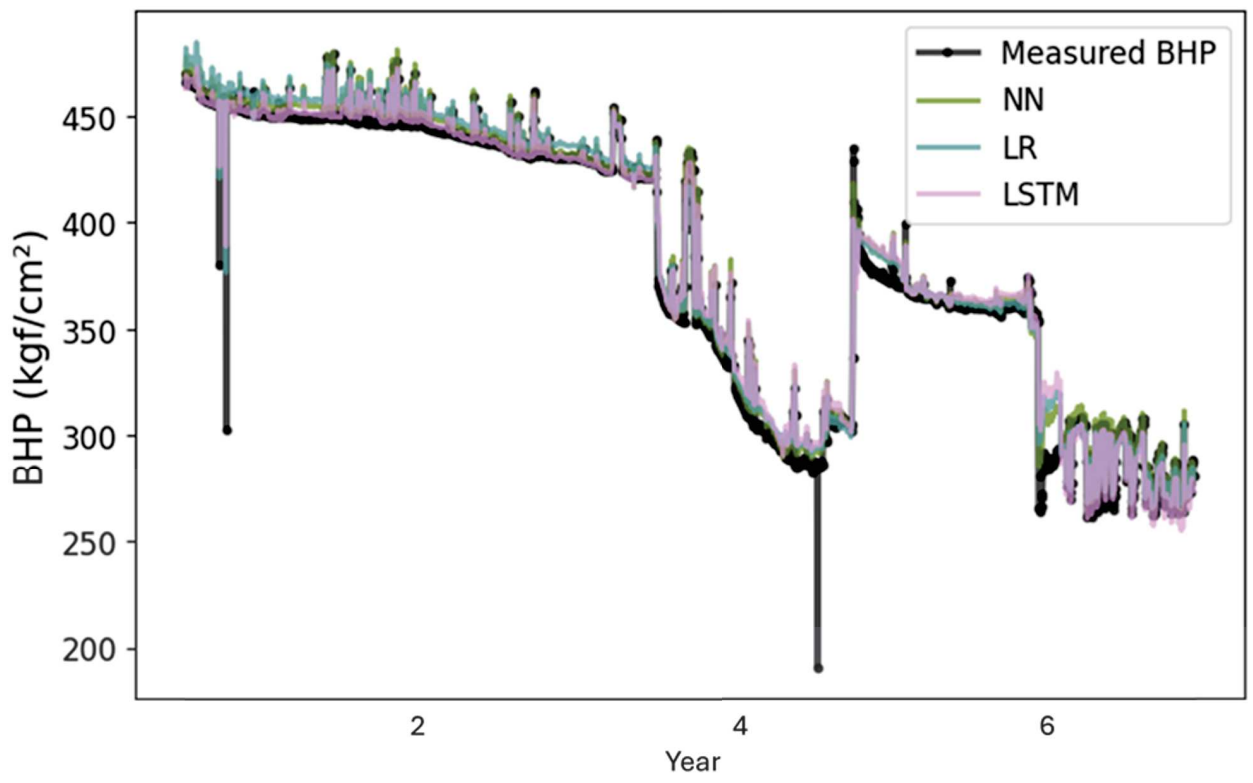

**Figure 13:** BHP estimates with selected models, applied to a well with long production history and abrupt changes in production behavior due to ICV adjustments (Field 1).

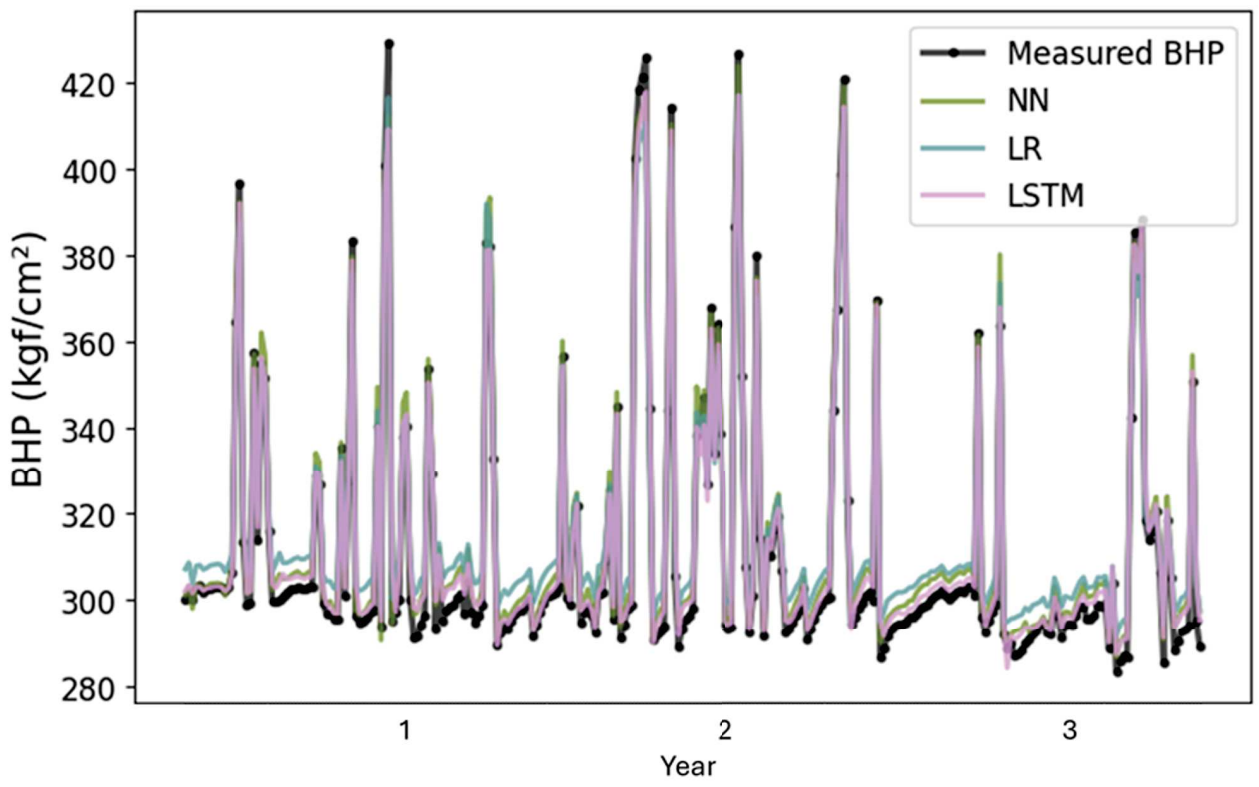

**Figure 14:** BHP estimates with selected models, applied to a well that experienced a significant increase in GOR over time and numerous production control adjustments (Field 1).

This observation is corroborated quantitatively in Figure 15, which shows the mean MAPE stratified by BHP range across both test datasets. The LSTM presents a clear advantage in the lowest BHP ranges (244–343 kgf/cm²), where errors from the linear regression are consistently up to 3 percentage points higher across multiple training runs, confirming the robustness of this finding. These low-pressure conditions, though less frequent, correspond to wells operating under significant restrictions or unusual flow regimes, precisely the scenarios where accurate BHP estimation is more challenging and can be critical. In the dominant operating range (343–441 kgf/cm²), all models perform comparably, with the linear regression showing a marginal advantage consistent with its lower variance. Overall, the linear regression remains a reliable choice for routine monitoring, while the LSTM is better suited for wells operating under atypical conditions.

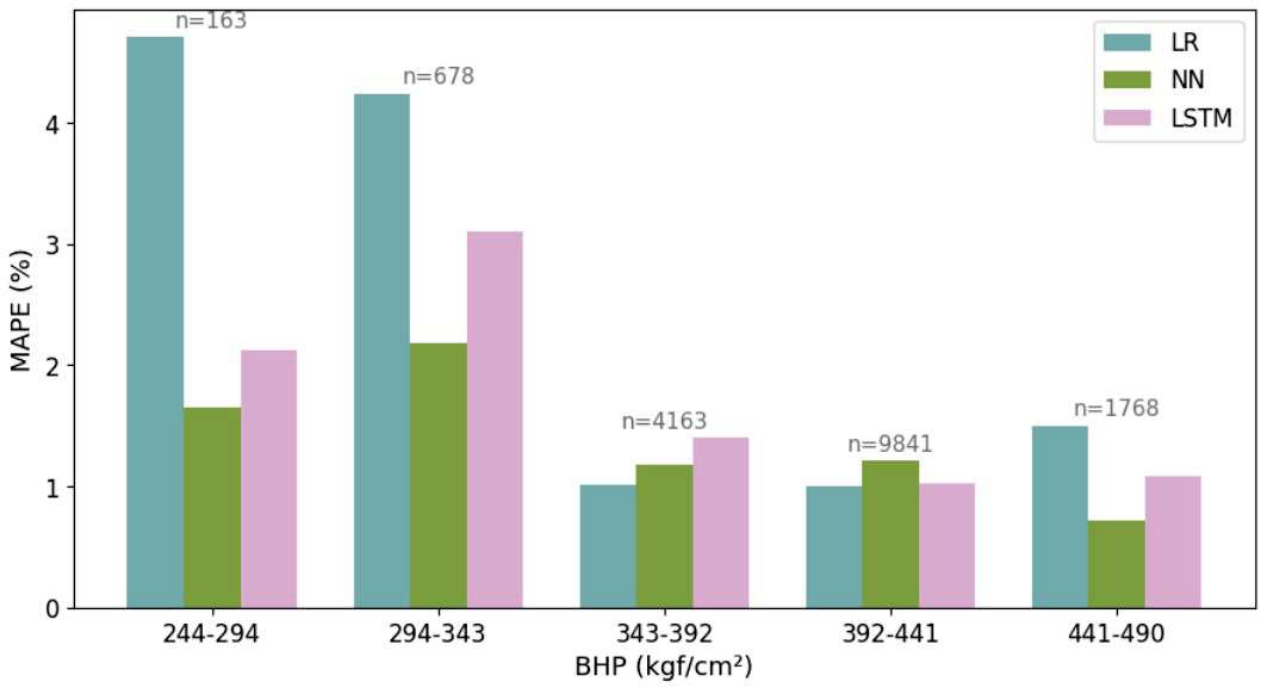

**Figure 15:** Mean MAPE stratified by BHP range for LR, NN, and LSTM models, computed across both blind test datasets for Field 1. Sample counts per bin are indicated above each group.

The linear regression, additionally, has the advantage of a training process more repeatable, with no noticeable variance. It can also be easily interpreted by observing the weights attributed to each of the input variables. The LSTM, by contrast, has internal representations that are not directly interpretable, which is a known limitation of deep learning methods and a subject of ongoing research in the field.

Crucially, our findings underscore the fundamental relationship between model performance and training data quality. As we stated in our previous work [34], robustness in the estimates requires datasets that adequately represent the full range of operational scenarios. The current study provides empirical evidence that expanded datasets significantly improve model robustness, yielding more consistent predictions with reduced uncertainty margins. This dataset-size effect proves particularly valuable for machine learning applications where data diversity directly impacts generalization capability.

### 4.3. Evaluating Transfer Learning for Field 2

In the final phase of our study, we investigated the application of transfer learning strategies for developing soft sensor replacements in scenarios involving new domains (fields or well groups) with insufficient training data for data-intensive supervised learning methods. We evaluated our top-performing LSTM and NN models using the two transfer learning approaches we described previously: (1) adding a new hidden layer to the original model while keeping pre-trained weights frozen, and (2) fine-tuning the pre-trained model. Both approaches were tested using data from one platform in Field 2 for retraining, with validation performed on two test sets: one containing data from a second platform (excluding the most recent year) and another comprising the final year's data from both platforms.

Notably, while our Field 1 dataset didn't include wells operating with gas lift systems, we incorporated this variable for Field 2, allowing better estimates for the wells that use it. This capability to accommodate new input variables is also one of the benefits of the transfer learning framework.

The base models were trained using the Adam optimizer (lr = 0.001) with early stopping. The LSTM architecture consists of a single recurrent layer with tanh activation, followed by dropout and a linear output layer, trained for up to 30 epochs. The MLP follows a similar structure, with a single hidden layer of 128 neurons with ReLU activation and MaxNorm regularization, trained under the same conditions.

For transfer learning, both fine-tuning strategies used the same 15,269 target field samples after outlier removal. In the fine-tuning approach, all layers were unfrozen and retrained with a reduced learning rate — 0.0001 for the LSTM and 0.0002 for the MLP — for up to 20 and 10 additional epochs, respectively. The lower learning rate is key to preserving the general representations learned from Field 1 while allowing adaptation to the new domain. In the new layer approach, the base weights were kept frozen and only the added layer was trained.

The results presented in Tables 10 and 11 compare error metrics between transfer learning implementations and models trained exclusively on Field 2 data, for both datasets. The metrics demonstrate excellent performance, confirming the methodology's applicability across different domains.

We can observe high errors from models applied directly to Field 2 without adaptation, as it presents a substantially different operational regime from Field 1: higher average BHP and wellhead pressures, greater water cuts and GOR values, and more frequent use of gas lift, a variable absent in Field 1. A model trained on Field 1 has never been exposed to the pressure dynamics associated with gas lift injection, which reduces fluid density and lowers BHP in a flow-regime-dependent manner. The MLP, lacking temporal memory, is particularly sensitive to this distributional mismatch and the ones in other variables, which explains its higher baseline error compared to the LSTM. The latter captures sequential patterns in pressure behavior that are partially shared across fields even when absolute values differ, what is consistent with its lower error on both test sets without transfer learning. These results reinforce both the necessity of domain adaptation and the complementary role of temporal modeling in bridging distributional gaps.

We obtained superior results with the fine-tuning approach, unfreezing all the layers and training with a lower learning rate. Figures 16 and 17 present the correlation between estimated and measured values for NN and LSTM models using this technique, compared against baseline cases without transfer learning. It is clear from the plots that both approaches can benefit from the transfer learning.

Finally, we plotted temporal BHP curves for two representative wells (Figure 18 and 19), comparing estimates from our best-performing LSTM model with and without the transfer learning against measured pressure data.

**Table 10**
Results obtained with representative models on the first test dataset, before and after transfer learning, for Field 2.

| **Model** | **Test Set 1** (11,229 samples) | | |
|---|---|---|---|
| | **MAPE** | **SMAPE** | **nRMSE** |
| NN1 *w/o TL* | 7.094 ±0.081 | 6.893 ±0.076 | 7.426 ±0.075 |
| NN1 *w/ New Layer* | 2.770 ±0.350 | 2.746 ±0.341 | 3.101 ±0.367 |
| NN1 *w/ Fine Tuning* | 2.733 ±0.086 | 2.766 ±0.085 | 3.542 ±0.107 |
| LSTM1 *w/o TL* | 3.409 ±0.128 | 3.588 ±0.138 | 5.877 ±0.308 |
| LSTM1 *w/ New Layer* | 2.737 ±0.346 | 2.808 ±0.376 | 3.960 ±0.620 |
| LSTM1 *w/ Fine Tuning* | **1.564 ±0.119** | **1.574 ±0.122** | **2.042 ±0.186** |

**Table 11**
Results obtained with representative models on the second test dataset, before and after transfer learning, for Field 2.

| **Model** | **Test Set 2** (2,077 samples) | | |
|---|---|---|---|
| | **MAPE** | **SMAPE** | **nRMSE** |
| NN1 *w/o TL* | 5.721 ±0.057 | 5.754 ±0.057 | 6.455 ±0.059 |
| NN1 *w/ New Layer* | 1.581 ±0.250 | 1.574 ±0.245 | 1.941 ±0.288 |
| NN1 *w/ Fine Tuning* | 1.143 ±0.053 | 1.144 ±0.053 | 1.435 ±0.052 |
| LSTM1 *w/o TL* | 1.000 ±0.153 | 1.006 ±0.156 | 1.209 ±0.177 |
| LSTM1 *w/ New Layer* | 1.178 ±0.136 | 1.169 ±0.133 | 1.660 ±0.154 |
| LSTM1 *w/ Fine Tuning* | **0.809 ±0.146** | **0.808 ±0.145** | **1.041 ±0.200** |

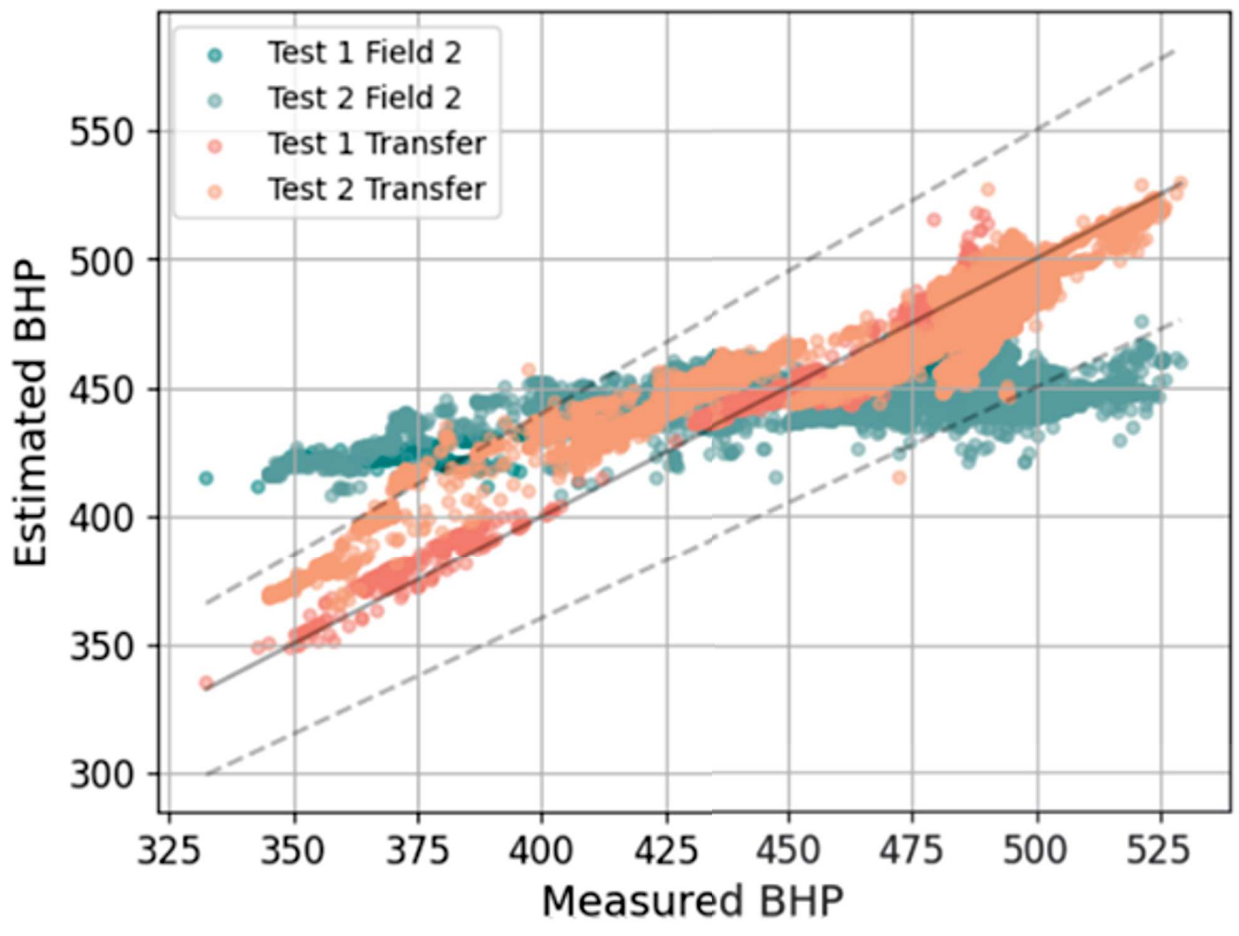

**Figure 16:** Correlation of estimated versus actual values for the BHP from our best NN model before and after transfer learning (Field 2). Pressure in kgf/cm².

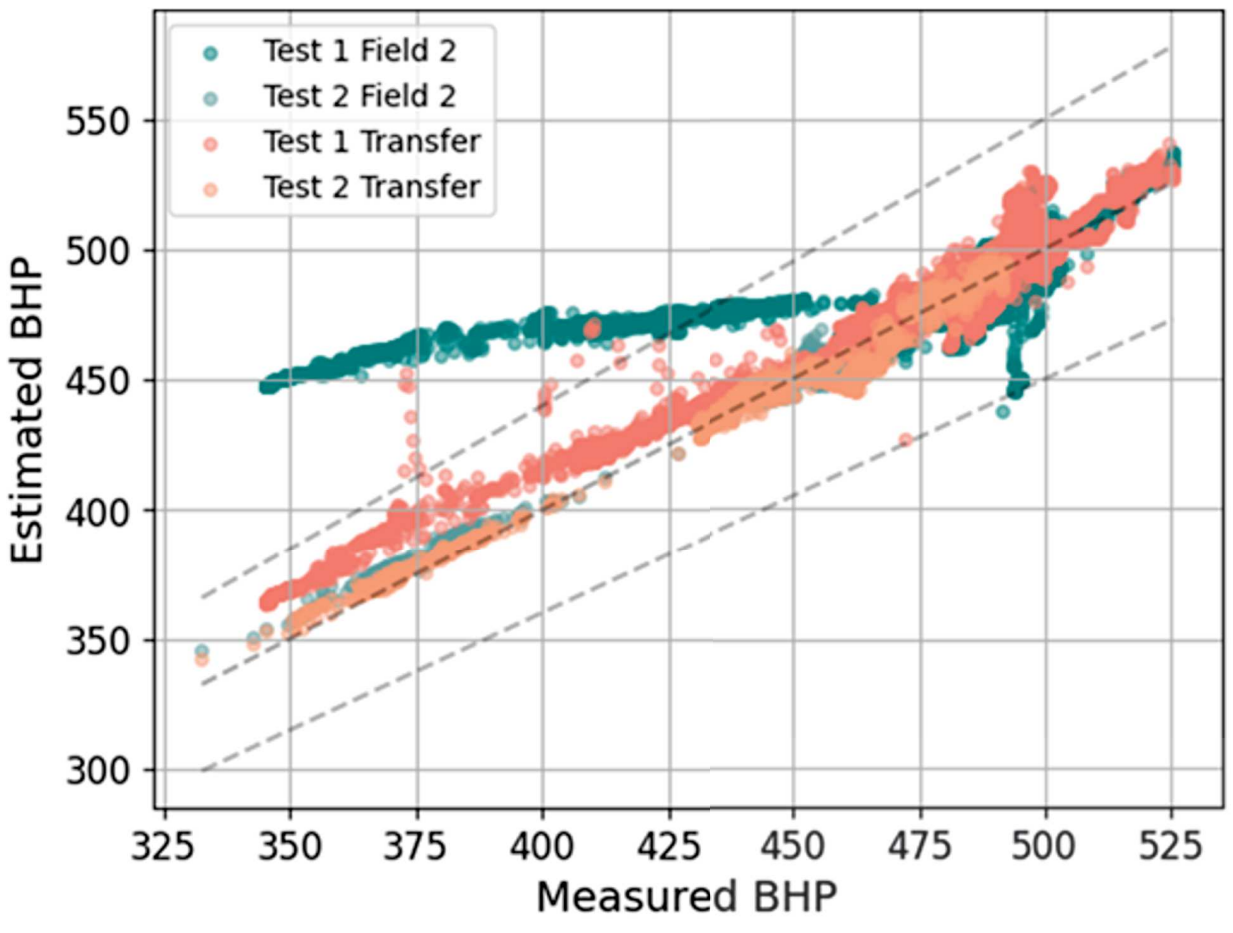

**Figure 17:** Correlation of estimated versus actual values for the BHP from our best LSTM model before and after transfer learning (Field 2). Pressure in kgf/cm².

In Figure 18, we bring a well with a long production history, which underwent significant changes such as major depletion and increases in GOR and water cut. The estimation remains highly accurate throughout the entire history. In the example shown in Figure 19, we have a recently drilled well with a history of high flow rates and significant restrictions in its opening. Also in this case, the BHP estimation maintained an error below 2% throughout the analyzed history. These results further validate both the quality of our methodology and its applicability to diverse operational scenarios.

Regarding computational efficiency, LSTM models required substantially longer training times (1-2 minutes per configuration) when running on a multi-processor cluster, compared to MLP neural networks (0.4 s) and regression models (0.075 s). However, all trained models achieved inference times of just milliseconds. This performance enables wells without an available PDG to be seamlessly integrated with PI systems for continuous data tracking and analysis, making them fully suitable for real-time monitoring applications such as reservoir management, artificial lift optimization, inflow performance analysis, and early detection of well anomalies.

In production environments where concept drift is expected, such as wells undergoing reservoir depletion or other changes in production parameters, periodic model updates are recommended. Retraining cadence requires only the most recent labeled data as input, and a practical choice is to perform it every four to six months.

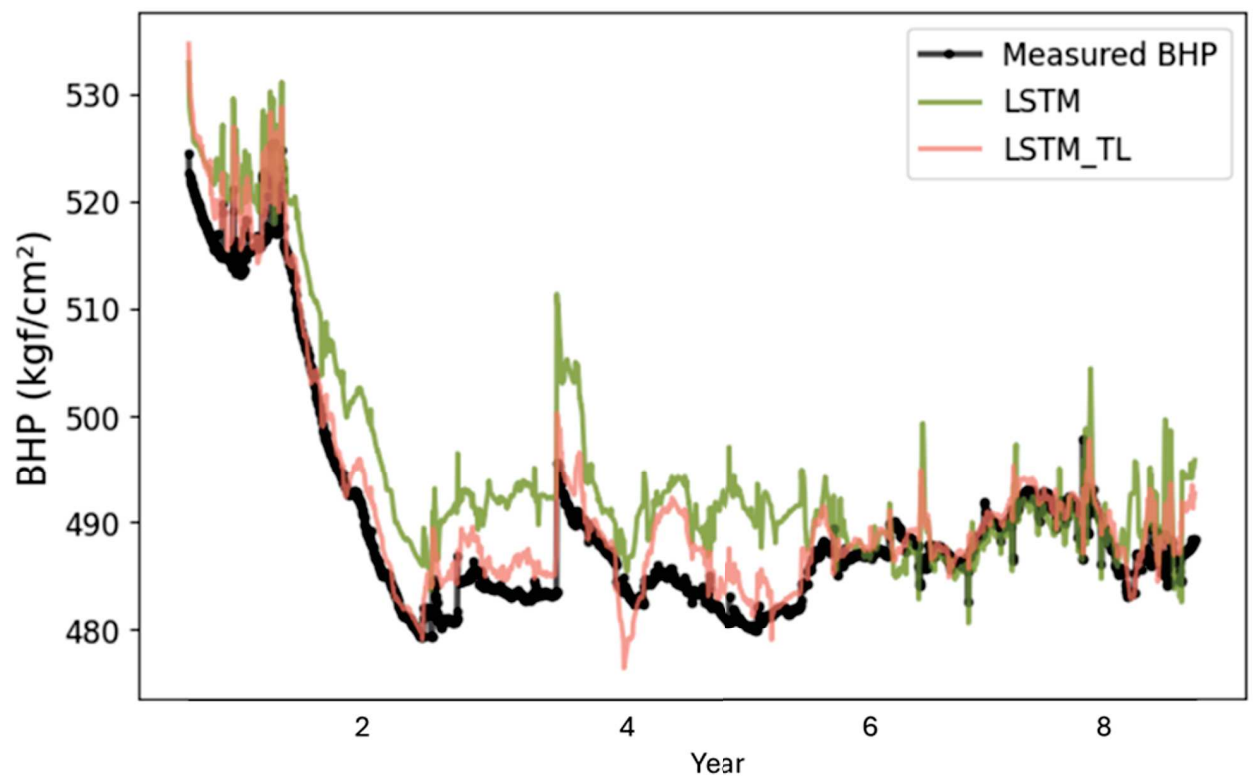

**Figure 18:** BHP estimates with the selected models, applied to a well with long production history (Field 2).

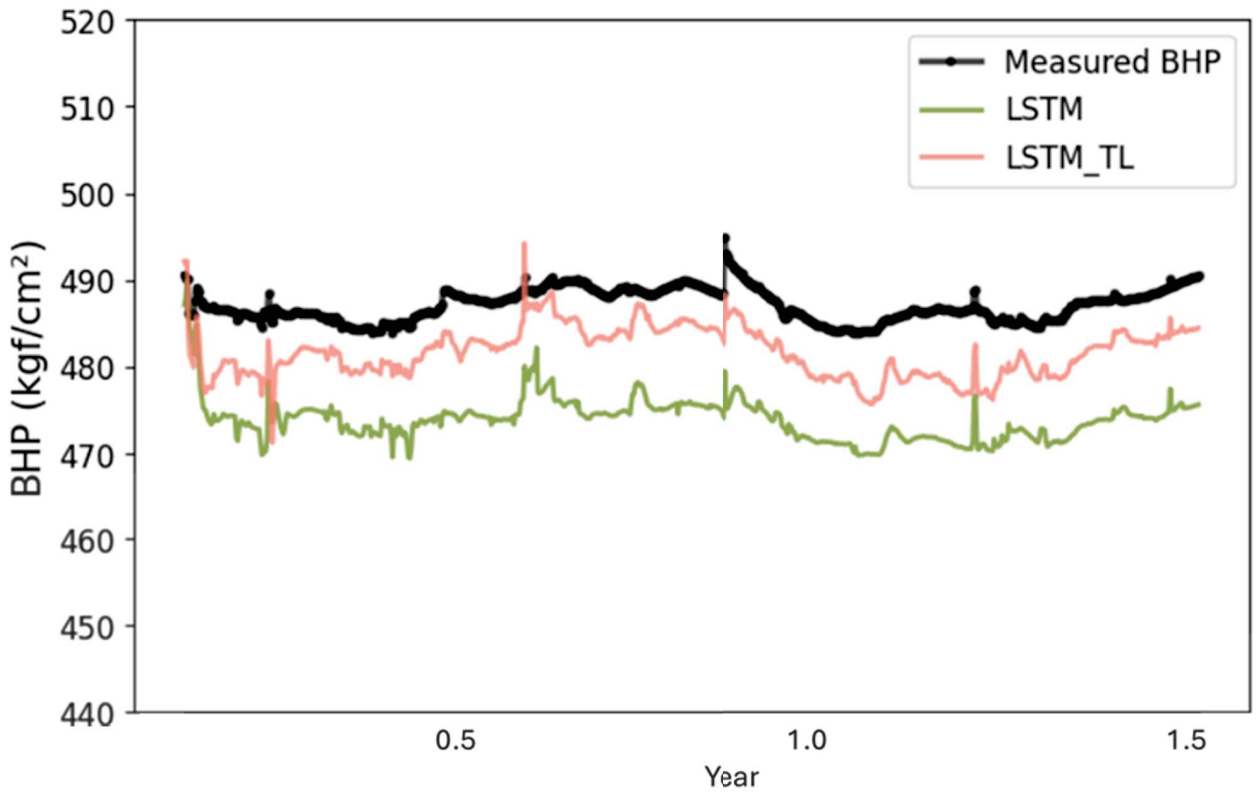

**Figure 19:** BHP estimates with the selected models, applied to a recently drilled well (Field 2).

The proposed methodology has some limitations worth noting. First, it is restricted to steady-state flowing BHP: samples with less than 2 hours of continuous production per day were excluded during preprocessing, and the model is not designed to capture transient behavior during well startup or shut-in events. Second, production rates are allocated daily from platform-level measurements based on the most recent well test, a process that introduces inherent uncertainty that propagates to model inputs and may affect accuracy during periods of significant well behavior changes. Third, although the framework is adaptable by design, as demonstrated by the transfer learning application across fields with distinct characteristics, it was developed and validated exclusively on offshore Pre-salt wells; applicability to onshore fields, heavy oil, ESP-lifted wells, or HPHT environments has not been assessed. Finally, as a purely data-driven approach, model performance depends on the availability of representative training data, and wells operating under conditions significantly outside the training distribution may require dedicated fine-tuning before deployment.

## 5. Conclusions

This work presented a practical application of machine learning in the oil and gas industry, deploying deep learning in the form of Long Short-Term Memory and transfer learning techniques to obtain estimators for the bottom-hole pressure in oil wells.

Our approach is purely data-driven, using as input variables the pressures and temperatures measured at the wellheads and on the platform, as well as the fluid production rates. To estimate the BHP, we used as a flagship model the LSTM, and as benchmarks MLP neural networks and regularized (ridge) linear regression.

We evaluated our framework in real-world scenarios, using data from a group of offshore platforms operating in the Brazilian pre-salt basin. First, we applied it to wells producing from a single giant oilfield, from which we obtained small errors, with MAPE $<1.5\%$ in our blind tests for a broad range of production conditions under regular flow. Although simpler models are already capable of delivering good results in most situations, the model with temporal dependency proved to be more efficient in more unusual situations and in future use simulations, which justifies its application despite its greater complexity and computational cost. A stratified error analysis by BHP range further confirmed that the LSTM advantage is most pronounced under atypical operating conditions, precisely where accurate estimation is most critical.

Subsequently, we evaluated transfer learning strategies, using the model trained on the first dataset as a base model for a second oilfield that had a smaller and less diverse history, also including a new variable, the gas lift rate required for some wells. We observed that the approach of fine-tuning using new data was the most effective and produced errors of the same order of magnitude as our first case. The domain shift between the two fields was formally quantified via symmetric KL divergence, providing a physically interpretable justification for the transfer learning framework.

With that, we demonstrated the applicability of the proposed methodology to diverse practical situations in oil and gas production, including important features such as adaptability to new domains and robustness to concept drift. The idea of a soft sensor for bottom-hole measurements can therefore be useful for offline studies or real-time monitoring (given the availability of wellhead and topside sensor data). Such estimators can be used as backup or reference in well digital twins or effectively replacing physical gauges in situations where they fail or when their installation is technically or economically prohibitive. The data-driven estimators can thus be considered valuable alternatives to ensure the availability of essential data for production monitoring and reservoir management, aiming at safe and efficient operations, together with value maximization.

# 6.

## CRediT authorship contribution statement

**M. A. Fernandes:** Conceptualization, Data curation, Methodology, Investigation, Software, Formal analysis, Writing – original draft, Writing – review & editing. **E. Gildin:** Conceptualization, Methodology, Formal analysis, Writing – review & editing, Supervision. **M. A. Sampaio:** Conceptualization, Methodology, Formal analysis, Writing – review & editing, Supervision.